\documentclass[sigconf]{acmart}

\usepackage{subfigure}
\usepackage{multirow}
\usepackage{multicol}
\usepackage[linesnumbered,boxed,ruled,commentsnumbered]{algorithm2e}
\usepackage{enumitem}
\setlist[itemize]{leftmargin=0.4cm}
\usepackage{pifont}

\usepackage{xcolor}
\usepackage{color, colortbl}
\definecolor{yymgray1}{HTML}{DDDFE8}
\definecolor{yymgray2}{HTML}{F2F2F2}
\definecolor{yympurple}{HTML}{AA93B4}
\definecolor{yymblue}{HTML}{E5F2FC}
\definecolor{yymblue1}{rgb}{0.6, 0.85, 0.95}
\definecolor{yymorange}{HTML}{FDECEE}
\definecolor{yymgreen}{HTML}{e6faef}
\definecolor{yymgreen1}{rgb}{0.7, 0.85, 0.85}

\AtBeginDocument{%
  \providecommand\BibTeX{{%
    \normalfont B\kern-0.5em{\scshape i\kern-0.25em b}\kern-0.8em\TeX}}}

 \copyrightyear{2025} 
\acmYear{2025} 
\setcopyright{acmlicensed}\acmConference[KDD '25]{Proceedings of the 31st ACM
 SIGKDD Conference on Knowledge Discovery and Data Mining V.2}{August 3--7,
2025}{Toronto, ON, Canada}
 \acmBooktitle{Proceedings of the 31st ACM SIGKDD Conference on Knowledge
 Discovery and Data Mining V.2 (KDD '25), August 3--7, 2025, Toronto, ON, Canada}
 \acmDOI{10.1145/3711896.3736912}
 \acmISBN{979-8-4007-1454-2/2025/08}

\begin{document}

\title{Discrepancy-Aware Graph Mask Auto-Encoder}

\author{Ziyu Zheng}
\orcid{https://orcid.org/0009-0000-3662-0832}
\affiliation{%
  \department{School of Computer Science and Technology,}
  \institution{Xidian University,}
  \city{Xi'an}
  \country{China}
}
\email{zhengziyu@stu.xidian.edu.cn}

\author{Yaming Yang}
\orcid{https://orcid.org/0000-0002-8186-0648}
\affiliation{%
  \department{School of Computer Science and Technology,}
  \institution{Xidian University,}
  \city{Xi'an}
  \country{China}
}
\email{yym@xidian.edu.cn}

\author{Ziyu Guan}
\authornote{Corresponding Author}
\orcid{https://orcid.org/0000-0003-2413-4698}
\affiliation{%
  \department{School of Computer Science and Technology,}
  \institution{Xidian University,}
  \city{Xi'an}
  \country{China}
}
\email{ziyuguan@xidian.edu.cn}

\author{Wei Zhao}
\orcid{https://orcid.org/0000-0002-9767-1323}
\affiliation{%
  \department{School of Computer Science and Technology,}
  \institution{Xidian University,}
  \city{Xi'an}
  \country{China}
}
\email{ywzhao@mail.xidian.edu.cn}

\author{Weigang Lu}
\orcid{https://orcid.org/0000-0003-4855-7070}
\affiliation{%
  \department{School of Computer Science and Technology,}
  \institution{Xidian University,}
  \city{Xi'an}
  \country{China}
}
\email{wglu@stu.xidian.edu.cn}


\begin{abstract}
Masked Graph Auto-Encoder, a powerful graph self-supervised training paradigm, has recently shown superior performance in graph representation learning. Existing works typically rely on node contextual information to recover the masked information. However, they fail to generalize well to heterophilic graphs where connected nodes may be not similar, because they focus only on capturing the neighborhood information and ignoring the discrepancy information between different nodes, resulting in indistinguishable node representations. In this paper, to address this issue, we propose a Discrepancy-Aware Graph Mask Auto-Encoder (DGMAE). It obtains more distinguishable node representations by reconstructing the discrepancy information of neighboring nodes during the masking process. We conduct extensive experiments on 17 widely-used benchmark datasets. The results show that our DGMAE can effectively preserve the discrepancies of nodes in low-dimensional space. Moreover, DGMAE significantly outperforms state-of-the-art graph self-supervised learning methods on three graph analytic including tasks node classification, node clustering, and graph classification, demonstrating its remarkable superiority. The code of DGMAE is available at \url{https://github.com/zhengziyu77/DGMAE}.
\end{abstract}



\maketitle

\section{Introduction}
\label{section:1}
Masked Auto-Encoders~\cite{MAE} have recently emerged as a prevalent technique for Self-Supervised Learning (SSL) methods to learn generic representations from large amounts of unlabelled data by reconstructing the masked portion of the data. Graph Masked Auto-Encoder (GMAE) is the extension of this category of approach to the domain of graph-structured data~\cite{graphmae, maskgae, gpt-gnn, s2gae}. They first corrupt the original graph by masking operations and then leverage the context of the nodes to recover the masked information. They are good at capturing the nuanced neighborhood information of a graph and have a robust capability to model the local structure of the nodes~\cite{rare}. Furthermore, they can learn more generalizable node representations that are suitable for a variety of downstream tasks, such as node classification~\cite{gcn, node,skipnode,AGMixup}, node clustering~\cite{cluster, HSAN,SCGC, Dink_net}, and graph classification~\cite{gin,gcla,san}, etc.

\begin{figure}[ht]
\centering
\includegraphics[width=0.95\columnwidth]{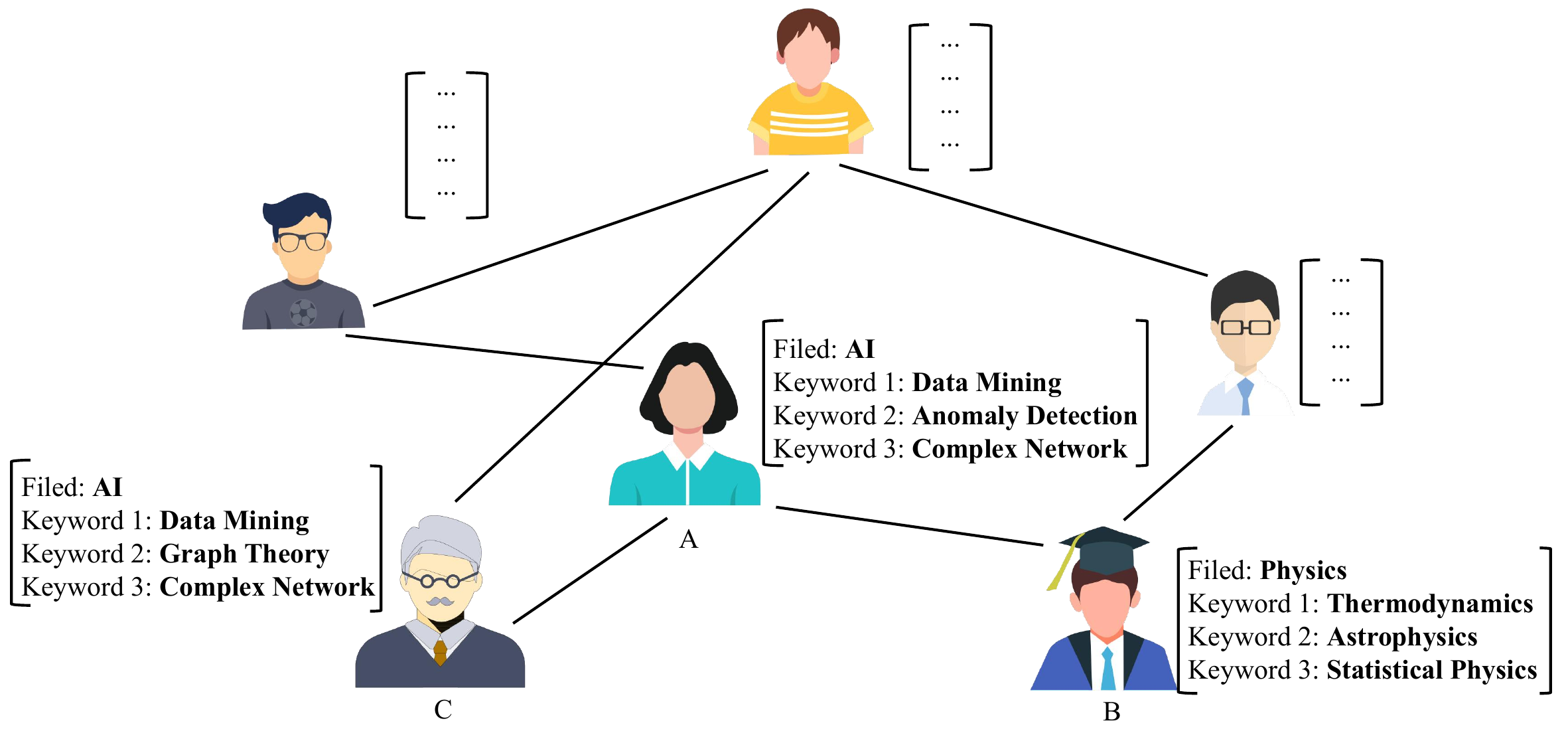}
\caption{The nodes represent scholars and edges represent the collaboration, with scholars in the network coming from various fields. When A is masked, C can provide the relevant context to recover features, while B may interfere with reconstruction.}
\label{fig.motivation2}
\Description{..}
\end{figure}

\begin{figure}[t]
\includegraphics[width=0.8\columnwidth]{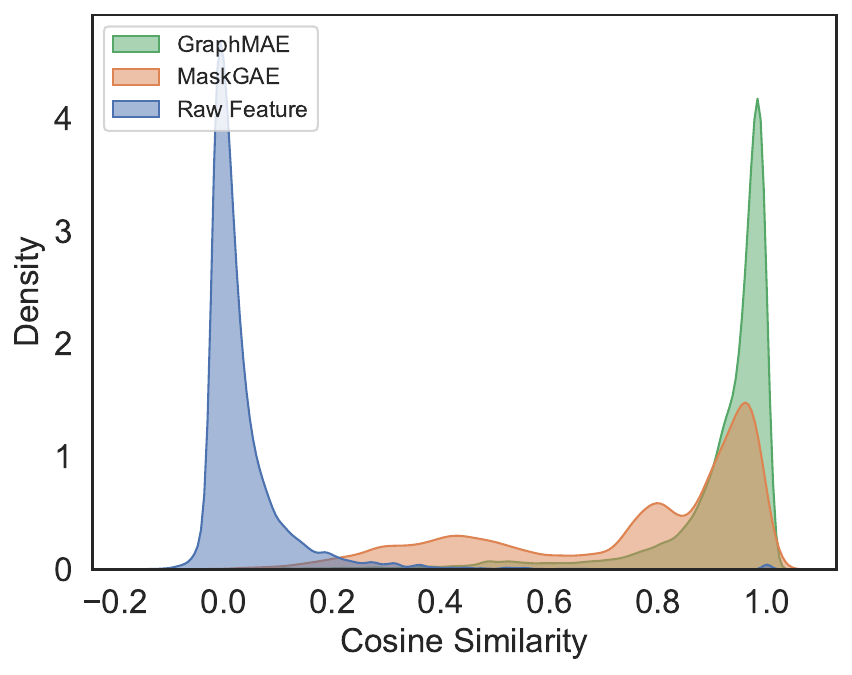}
\caption{The effect of GMAE methods on the similarity of the representations of nodes with different classes on the Squirrel.}
\label{fig.motivation1}
\Description{..}
\end{figure}

Existing GMAE methods typically leverage the relevant information between the target node and its masked context to reconstruct the original information. Unfortunately, existing models lack awareness of the discrepancy information between nodes, causing the reconstructed representation to deviate from the original semantics. For example, as a toy scholar collaboration network depicted in Figure \ref{fig.motivation2}, scholars A and B have a collaboration relationship, while they are from different research fields and their attributes are highly discrepant. Assume that we mask scholar A and let an existing GMAE model reconstruct its feature based on its neighborhood information, scholar B's feature will critically interfere with the reconstruction.

Theoretically, previous studies~\cite{augmae,u-mae} have proved that MAEs essentially align positive context pairs in an implicit way. In the context of GMAE, studies~\cite{graphmae,augmae} perform feature reconstruction via node masks to align the context of masked nodes, and studies~\cite{maskgae, s2gae} perform link reconstruction via edge masks to align the context of connected nodes. In other words, the alignment property of GMAE will drive connected nodes to have similar reconstructed features. However, in practice, real-world networks are very likely to be heterophilic~\cite{ud-gnn}, i.e., connected nodes may belong to different classes. This contradiction between theory and practice will make it difficult for GMAE methods to distinguish nodes with different classes.

To further verify this issue more intuitively, we conduct a small experiment to study the effect of GMAE methods on the similarity of connected nodes of different classes. The result is shown in Figure~\ref{fig.motivation1}. As we can see, compared with the similarity of the orignal features, two representative traditional GMAE methods significantly enhance the similarity of heterophilic node pairs. This exactly explains why current GMAE approaches are less effective on heterophilic than homophilic graphs.


In this work, we aim to address this issue by explicitly leveraging the discrepancy information to help the model learn more discriminative node representations. The main idea is to preserve the discrepancies between nodes in the low-dimensional embedding space. However, we are facing two aspects of challenges: (1) How to compute the discrepancy information between nodes in the graph domain? Due to the lack of supervised information, nodes cannot perceive which neighbors are more helpful for feature reconstruction. (2) How to preserve the discrepancy information of low-dimensional embedding representations? We cannot directly compute the discrepancies between nodes in the embedding space since some nodes are masked.

In response to the two challenges above, we propose a new graph generative learning method called Discrepancy-Aware Graph Mask Auto-Encoder (DGMAE). Its core innovation lies in redefining the self-supervised paradigm—shifting from "recovering consistency" to "preserving discrepancy", which can effectively deal with both homophilic and heterophilic graphs. For the first challenge, a difference operator is developed to compute the discrepancy information between nodes. To pay more attention to the information of node pairs with large discrepancies, an edge sampling strategy based on encoded attention is introduced to preserve these important discrepancies. For the second challenge, we leverage the context discrepancies to reconstruct the original feature discrepancies and perform joint training with the original feature reconstruction. Our main contributions can be summarized as follows:

(1) We propose a new graph generative learning model dubbed DGMAE. It no more only focuses on the consistency of nodes with their contextual neighbourhood, explicitly preserves the discrepancies of nodes in low-dimensional representations, which can help the model learn more discriminative node representations.

(2) We designed an adaptive discrepancy selection module. It can select the optimal discrepancy information based on the graph attention weights which are shared by the GNN encoder. This is more flexible and can help the model adapt to graphs with different homophilic ratios.

(3) We conduct extensive experiments on 17 graph benchmark datasets involving both homophilic and heterophilic graphs. The results show that DGMAE can outperform state-of-the-art baselines on various downstream tasks including node classification, node clustering, and graph classification.

\begin{figure*}[ht]
\centering   
\includegraphics[width=0.96\linewidth]{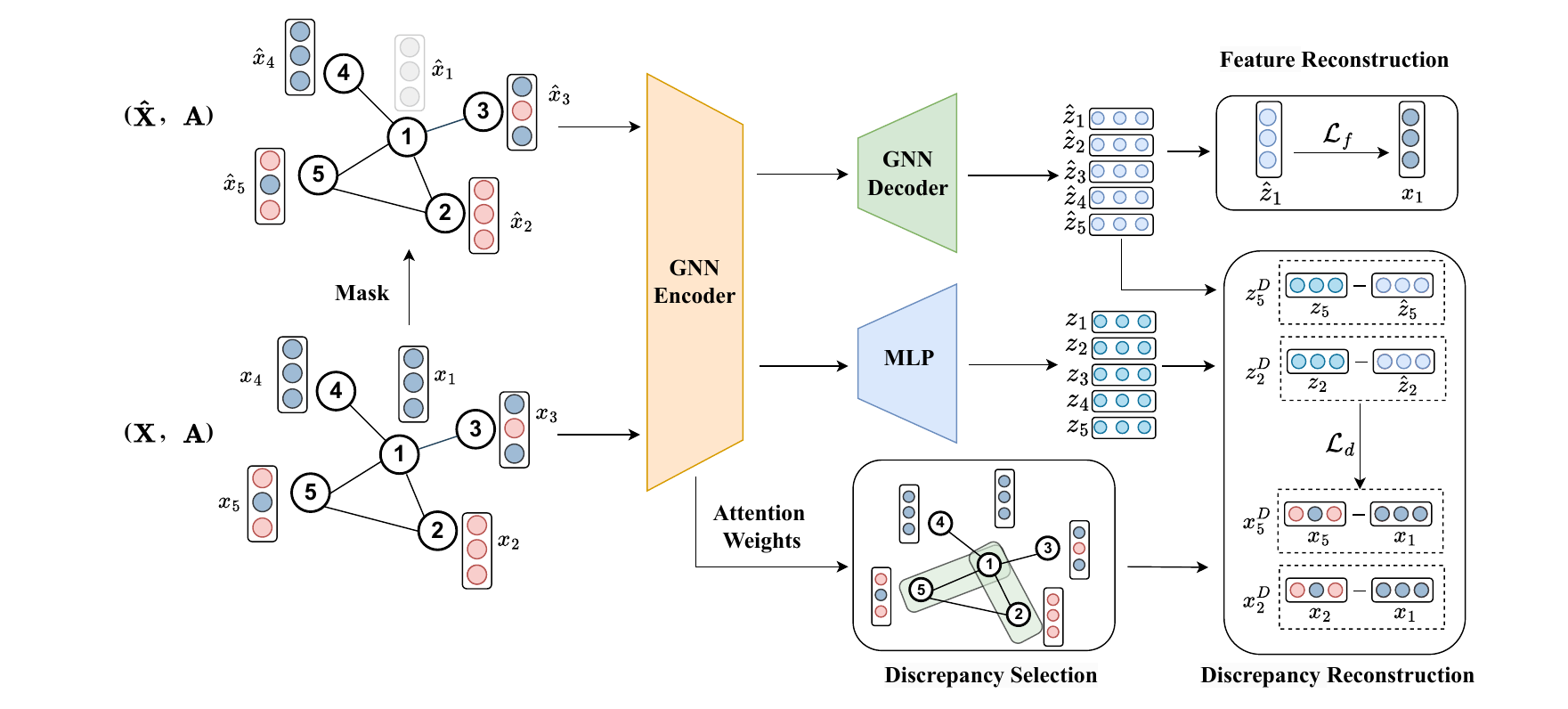} 
\caption{Illustration of the DGMAE framework, which consists of three components: (1) discrepancy predefinition, (2) feature reconstruction, and (3) discrepancy reconstruction. The feature reconstruction channel inputs the masked version of the node feature $X$ and the original adjacency matrix $A$ and outputs the reconstruction of the feature. The discrepancy reconstruction channel shares the same encoder and mitigates the problem of node confusion by learning the discrepancies.} 
\label{Fig.model}
   \Description{..}
\end{figure*}

\section{Related Work}
\subsection{Masked Genaratived Methods}
Generative graph learning methods~\cite{vgae,arvga} learn node representations by reconstructing features or structures. Among them, masked generative learning methods learn node representations by masking nodes or edges and reconstructing them. GraphMAE~\cite{graphmae} reconstructs the original features from the visible information by masking the node features. MaskGAE~\cite{maskgae} learns the structural information by reconstructing the edges and paths using path masks. SimGOP~\cite{simgop} learn the consistent information of node-contexts. S2GAE \cite{s2gae} utilizes multilayer representations for cross-correlation decoding. AUG-MAE~\cite{augmae} introduces adaptive masks to learn the presence of hard positive samples. Bandana~\cite{bandana} replaces binary masks with non-discrete bandwidth masks based on MaskGAE~\cite{maskgae}. However, these methods rely on consistent information between nodes and their neighbours, which may inadvertently lead to over-similarity among node representations and hinder task performance in heterophilic graphs. Our approach leverages discrepancy information to foster more distinctive and expressive representations.

\subsection{GNNs for Heterophily Graphs}
Current Graph Neural Networks (GNNs) directly or indirectly introduces a high-pass filter in the encoding to extract heterophilic information. GPR-GNN~\cite{GPRGNN} processes the high-frequency and low-frequency graph signal utilizing a learnable signed weight. FAGCN~\cite{fagcn} and ACM-GCN~\cite{acmgnn} perform low-pass filtering and high-pass filtering for each node and adaptively fuse the node embeddings of each filter. GloGNN~\cite{glognn} aggregates global information to learn signed coefficient matrix to capture correlations between nodes. However, these methods are based on supervised learning, where the weights of high-frequency information depend on the label loss. In contrast, we construct a learnable high-frequency signal not used as encoding input but as a reconstruction target to learn the discrepancy between nodes.

\subsection{SSL on Heterophilic Graphs}
Self-supervised learning (SSL) on heterophilic graphs has gained considerable interest, prompting the development of methods applicable to both homophilic and heterophilic graphs. Greet~\cite{greet} emploies low-pass and high-pass filters for two-view comparative learning. GraphACL~\cite{graphacl} leverages the notion that similar nodes exhibit analogous neighborhood distributions~\cite{homophily} to capture one-hop local neighborhood information and two-hop monophily similarity. It maximizes mutual information with neighborhood patterns, thereby implicitly aligning two-hop neighbors. NWR-GAE~\cite{NWR-GAE} recovers node neighborhoods by predicting feature distribution based on Wasserstein distance-based optimal transport loss. DSSL~\cite{dssl} models the generative process of nodes and links through latent variable semantics, decoupling distinct neighborhood latent semantics into a self-supervised framework. These methods rely on edge or neighborhood distributions to learn heterophilic information and require expensive time costs in edge sampling and neighborhood sampling. Our approach utilizes shared attention to perform adaptive differential selection to avoid complex sampling processes.

\section{Notations and Definitions}
A graph is defined as $G=(\mathcal{V},\mathcal{E})$, where $\mathcal{V}=\{v_1,v_2,\cdots,v_n\}$ represents the set of the nodes and $\mathcal{E} \subseteq \mathcal{V} \times \mathcal{V}$ represents the edge set. Let $N (i)$ denote the neighbor set of node $i$. Let $X \in \mathbb{R}^{n\times d}$ denote the feature matrix, where $d$ is the dimensionality. Let $A \in \mathbb{R}^{n \times n}$ denote the adjacency matrix. Its element $A_{ij}=1$ if there exists an edge between node $i$ and $j$, and otherwise, $A_{ij}=0$. The symmetric normalized adjacency matrix is denoted as $\widetilde{A} = D^{-1/2}AD^{-1/2}$, where $D$ is the diagonal degree matrix, and $D_{ii} = \sum_{j}A_{ij}$. The symmetric normalized graph Laplacian matrix is denoted as $L_{sym} = I-\widetilde{A}$. 

\section{Methodology}

\subsection{Overview}
In this section, we describe DGMAE in detail. The key idea is to learn discrepancy information by discrepancy feature reconstruction, which encourages the model to preserve the original discrepancies in low-dimensional representation space. Figure \ref{Fig.model} shows an overview of the model architecture. There are two main branches, one branch is to reconstruct the original features and the other branch is to reconstruct the feature discrepancy.

\subsection{Original Feature Reconstruction}
We focus on the self-supervised learning on graphs in a generative learning manner. Following most previous generative learning methods~\cite{graphmae,augmae,s2gae,maskgae,graphmae2}, we adopt the widely used masked GAE as the backbone model of our DGMAE. 
Specifically, we first randomly sample a subset of nodes as the masked nodes under the Bernoulli distribution, denoted as $\widehat{V}$. The features of masked nodes are set to $0$. Thus, the masked node features can be denoted as:
\begin{equation}
    \widehat{x}=\left\{
    \begin{aligned}
    0, v_i \in \widehat{V}, \\
    x_i, v_i \notin \widehat{V}.
    \end{aligned}
    \right.
\end{equation}
The objective of feature reconstruction is to reconstruct the features of the mask nodes. By exploiting the message passing and aggregation in GNNs, each node can obtain the relevant information in neighbors to enhance the node representation. It is expected to recover the original features of the masked node through context representations. Given an encoder $f_E$ and a decoder $f_D$, $f_E$ encodes the masked features into a low-dimensional space, $f_D$ reconstructs masked node information by utilizing the unmasked context: 
\begin{equation}
\label{eq-unmask}
    \hat{H} = f_E(\hat{X}, A), \quad \hat{Z}=f_D(\hat{H}, A).
\end{equation}
$\hat{H}$ denotes the encoded node representation and $\hat{Z}$ denotes the decoded representation, we reconstruct the features by the Scaled Cosine Error (SCE) loss~\cite{graphmae,graphmae2,augmae,rare}:
\begin{equation}
\label{eq-ft-re}
    \mathcal{L}_f = \frac{1}{|\widehat{V}|} \sum \limits_{i \in \widehat{V}} (1-⟨\mathbf{z_i} ,\mathbf{x_i}⟩)^{\gamma_1},\gamma_1 > 1,
\end{equation}
where $\gamma_1$ is the scaling coefficient and $⟨\mathbf{z_i^D},\mathbf{x_i^D}⟩=\frac{x_i^T \cdot \hat{z}_i}{\Vert x_i^T \Vert \cdot \Vert \hat{z}_i \Vert}$. The decoding feature is essentially the contextual information obtained by aggregating the neighborhood. The feature reconstruction predicts the original features by learning the contextual representations of the mask nodes. However, in heterophilic graphs, their contexts lack enough relevant information to recover the original features. Minimizing the discrepancy between the contextual representation and the original features may destroy the primitive node semantic, leading to sub-optimal node representations, as shown in Figure~\ref{fig.motivation1}. To this end, we proposed the feature discrepancy reconstruction module.

\subsection{Feature Discrepancy Reconstruction}
In this work, we introduce a new reconstruction task that captures the discrepancy information between the masked node and its neighbors, which helps the model learn more discriminative node representations. The goal of this task is to reconstruct the feature discrepancies between nodes.

\subsubsection{Predefined Raw Feature Discrepancies} 
To preserve the original discrepancy information in the low-dimensional representation space, we treat the discrepancies of original features as target reconstruction signals. Thus the discrepancy between a node and its neighbor nodes is computed as follows:
\begin{equation}
x_i^{D} = x_i-\sum\limits_{j \in \mathcal{N}(i)}\frac{1}{\sqrt{d_i} \sqrt{d_j}}x_j.\label{Eq4}
\end{equation}
To express the discrepancy between the original features intuitively, we adopt a simple feature subtraction as the feature discrepancy metric after normalizing the original features. It can be observed that the signal discrepancy between the node and its first-order neighbors is equivalent to applying the graph Laplace operator $L_{sym}$ to the original feature $X$, which can be denoted as:
\begin{equation}
    X^D = (I- \widetilde{A})X = L_{sym}X.
\end{equation}
$L_{sym}$ is computed as a normalized symmetric matrix with the weights of the edges denoted as $D^{-1/2}AD^{-1/2}$. It has been used in many studies of heterophilic graphs as a high-pass filter~\cite{fagcn, acmgnn} for the extraction of high-frequency information. The difference is that they~\cite{fagcn, acmgnn} directly strengthen the discrepancies between the node and its neighbors. In contrast, our work does not use them as direct inputs but treats them as supervised signals to preserve the original discrepancy information during training.

\subsubsection{Adaptive Discrepancy Selection}
Since different node pairs have different discrepancies between them, we proposed an adaptive selection mechanism to accommodate their personalities. Specifically, we use attention coefficients used in the encoder $f_E$ to guide the selection discrepancies for node pairs with larger discrepancies. The attention of each edge is computed as follows:
\begin{equation}
    w_{ij} = \frac{\exp(LeakyRelu( [W\hat{h_i} || W\hat{h_j]}))}{\sum_{k\in\mathcal{N}(i)}\exp(LeakyRelu([W\hat{h_i} || W\hat{h_k]}))},
\end{equation}
where $||$ denotes the concatenation operation, $\hat{h}_i$ and $\hat{h}_j$ are the embedding of node $v_i$ and $v_j$, $W$ is a learnable parameter matrix, and $w_{ij}$ denotes the representation similarity between node $i$ and node $j$. Here, since we aim to compute the discrepancy between node pairs, we subtract $w_{ij}$ from 1 and compute a sampling probability for each high-discrepancy node pair, described as follows:
\begin{equation}
    p_{ij} = min((1-w_{ij}) \cdot p_c, p_\tau),
\end{equation}
 where $p_c$ is the probability of removing edges, $p_\tau$ is the cut-off probability, $p_c$ controls the size of the edge sampling, $p_\tau$ to avoid destroying the graph structure due to over-sampling. Unlike in GCA~\cite{GCA} where fixed weights are used to guide edge sampling, here we guide edge sampling through a shared attention weight. When the attention weights between pairs of nodes are low, which means that nodes are more dissimilar and have a higher probability of being selected, we filter edges with greater discrepancy through a Bernoulli distribution: 

\begin{equation}
    m_{ij} \sim Bernoulli(p_{ij}),
\end{equation}
where $m_{ij}=1$ indicates that node $i$ and node $j$ are more likely to be dissimilar. Thus, Eq.\ref{Eq4} can be rewritten as follows:
\begin{equation}
    x_i^{D} = \sum\limits_{j \in \mathcal{N}(i)}\frac{m_{ij}}{\sqrt{d_i} \sqrt{d_j}}(x_i - x_j).\label{Eq9}
\end{equation}
Since $m_{ij}$ is computed based on the attention coefficients for feature reconstruction, when a node's neighbors are more important for the feature reconstruction process, the probability of them being selected for the discrepancy reconstruction will be lower.

\subsubsection{Features Discrepancy Reconstruction}
The measurement of the discrepancies between node pairs in a low-dimensional embedding space is difficult, as existing GMAE methods may easily result in homogeneous embeddings. Intuitively, in a masked graph, the reconstructed representation of a masked node relies on its unmasked neighbors, while for unmasked graphs, the corresponding masked node has a complete neighborhood. Hence, in this work, we treat the discrepancy in the contextual representation of the masked and unmasked graphs as the discrepancy information in the low-dimensional space. We have already computed the node representations of the masked graph, i.e., $\hat{Z}$, by Eq.~\ref{eq-unmask}. Here, we compute the node representations of the unmasked graph i.e., $Z$, as follows: 
\begin{equation}
    H = f_E(X,A), \quad Z =project_\theta(H).
\end{equation}
Here, the encoder $f_E$ is shared with the feature reconstruction module described by Eq.~\ref{eq-unmask}. A two-layer MLP is used to project the node representation to the same feature space as $\hat{Z}$. $\theta$ is a learnable parameter. Thus, the discrepancy in contextual representation in masked and unmasked graphs can be defined as follows:
\begin{equation}
    z^D_i = z_i - \hat{z}_i.
\end{equation}
During the feature reconstruction process,  we use GAT~\cite{gat} as a decoder to recover the original features. Therefore, the GAT decoder is able to extract the common information of the node and visible neighbors well to obtain the consistent representation. Subsequently, the discrepancy information is computed by subtracting the common information.
To preserve the discrepancy information between nodes, we utilize adaptively extracted original feature discrepancy as reconstructed signals and adopt the SCE loss used in Eq.~\ref{eq-ft-re} to reconstruct the original discrepancy:

\begin{equation}
\mathcal{L}_d =\frac{1}{|\hat{V}|} \sum \limits_{i \notin \hat{|V|}} (1- 
⟨\mathbf{z_i^D} ,\mathbf{x_i^D}⟩)^{\gamma_2}, \gamma_2 > 1 ,
 \label{eq.12}
\end{equation}
where $\gamma_2$ is the scaling coefficient. The nodes that perform discrepancy reconstruction select the set of unmasked nodes. Reconstructing the feature discrepancy between nodes and their neighbors can help nodes preserve their personal features, avoiding the generation of homogeneous node embeddings. 

\begin{table*}
\normalsize
    \centering
     \caption{Node classification accuracy (\%) on heterophilic graphs. The best and second-best results are highlighted in bold and underlined, respectively.}
    \resizebox{\linewidth}{!}{\begin{tabular}{c|cccccccc}
        \toprule
           Method & Texas & Cornell & Wisconsin & Chameleon & Squirrel & Crocodile & Actor &Roman \\
        \midrule
        DGI&58.53$\pm$2.98&45.33$\pm$6.11&55.21$\pm$1.02&60.27$\pm$0.70&26.44$\pm$1.12&51.25$\pm$0.51&28.30$\pm$0.76&63.71$\pm$0.63\\
        GCA& 52.92$\pm$0.46& 52.31$\pm$1.09&59.55$\pm$0.81&63.66$\pm$0.32&48.09$\pm$0.21&60.73$\pm$0.28&28.77$\pm$0.29&65.79$\pm$0.75\\
        CCA-SSG & 59.89$\pm$0.78 & 52.17$\pm$1.04 & 58.46$\pm$0.96 & 62.41$\pm$0.22 & 46.76$\pm$0.36 & 56.77$\pm$0.39&27.82$\pm$0.60& 67.35$\pm$0.61  \\
        BGRL & 52.77$\pm$1.98 & 50.33$\pm$2.29 & 51.23$\pm$1.17 &  64.86$\pm$0.63 & 36.22$\pm$1.97& 53.87$\pm$0.65&28.80$\pm$0.54 & 68.66$\pm$0.39  \\
        SP-GCL & 80.36$\pm$5.64 &\underline{78.33$\pm$4.26}  & 78.26$\pm$6.34 &  65.28$\pm$0.53 &  52.10$\pm$0.67&  61.72$\pm$0.21&28.94$\pm$0.69 & 70.88$\pm$0.35  \\
        GREET  & \underline{82.70$\pm$6.53} & 73.78$\pm$4.02 & \underline{82.94$\pm$4.96} & 63.64$\pm$1.26 & 42.29$\pm$1.43 & 67.28$\pm$0.45&\underline{36.55$\pm$1.01} & 50.69$\pm$0.34  \\
        GraphACL & 71.08$\pm$0.34 & 59.33$\pm$1.48 & 69.22$\pm$0.40 & 69.12$\pm$0.24 & 54.05$\pm$0.13 & 66.17$\pm$0.24 &30.03$\pm$0.13& \underline{74.91$\pm$0.28}\\
        
        \midrule
        VGAE&50.27$\pm$2.21&48.73$\pm$4.19&55.67$\pm$1.37&42.65$\pm$1.27&29.13$\pm$1.16&45.72$\pm$1.53&26.99$\pm$1.56&50.89$\pm$0.96\\
        DSSL & 62.11$\pm$1.53  & 53.15$\pm$1.28 & 62.25$\pm$0.55 & 66.15$\pm$0.32 &40.51$\pm$0.38 & 62.98$\pm$0.51& 28.15$\pm$0.31&71.70$\pm$0.54\\
        GraphMAE & 55.14$\pm$6.75 & 51.62$\pm$6.45 & 55.69$\pm$3.06 & 68.67$\pm$1.61 &44.91$\pm$1.94 & 67.74$\pm$0.95&25.61$\pm$1.22 &46.73$\pm$0.55  \\
        SeeGera & 68.10$\pm$7.52& 64.05$\pm$4.84 &57.64$\pm$4.40& 48.50$\pm$2.58 & 33.98$\pm$0.92 & 30.73$\pm$0.77 & 28.98$\pm$1.70 & 30.73 $\pm$ 0.77 \\
        MaskGAE & 64.86$\pm$1.27&43.24$\pm$5.10&46.67$\pm$3.31 & 61.36$\pm$0.88&50.40$\pm$0.80&67.24$\pm$0.45&26.32$\pm$0.59&49.98$\pm$0.59  \\

        NWR-GAE & 69.62$\pm$6.66  & 58.64$\pm$5.61 & 68.23$\pm$6.11 & 72.04$\pm$2.59&\underline{64.81$\pm$1.83} & \underline{72.68$\pm$0.90}& 30.17$\pm$0.17&46.35$\pm$0.48\\
        AUG-MAE & 72.97$\pm$7.15& 65.14$\pm$6.22 & 76.47$\pm$4.11 & \underline{72.19$\pm$1.55} & 53.23$\pm$1.62& 70.15$\pm$0.32& 26.36 $\pm$ 0.92& 51.13$\pm$0.29 \\
        Bandana & 68.73$\pm$ 4.24 & 46.84$\pm$3.23 & 61.76$\pm$2.31 & 50.72$\pm$1.89 &35.70 $\pm$0.33 & 68.25$\pm$0.63&28.32$\pm$0.94 &21.35 $\pm$ 0.49 \\
        \textbf{DGMAE} & \textbf{88.11$\pm$5.16} & \textbf{78.65$\pm$4.59} & \textbf{88.43$\pm$3.56} & \textbf{75.50$\pm$1.17} & \textbf{72.47$\pm$1.77} & \textbf{75.48$\pm$0.50} & \textbf{36.61$\pm$0.74} & \textbf{76.62$\pm$0.41} \\
             
        \bottomrule
    \end{tabular}}
    \label{table.1}
\end{table*}

\begin{table*}
\Huge
    \centering
        \caption{Node classification accuracy (\%) on homophilic graphs. The best and second-best results are highlighted in bold and underlined, respectively.}
    \resizebox{\linewidth}{!}{\begin{tabular}{c|ccccccccc}
        \toprule
        Method  & Cora & Citeseer & Pubmed &  Computer & Photo & Physical & CS & WikiCS & Flicker\\
        \midrule
        DGI &  82.30$\pm$0.60&71.80$\pm$0.70 & 76.8$\pm$0.6&85.10$\pm$0.55&91.42$\pm$0.35&94.50$\pm$0.23&91.92$\pm$0.30&75.79$\pm$0.16&45.10$\pm$0.22 \\
        BGRL & 81.10$\pm$0.15 & 71.22$\pm$0.70 &  79.99$\pm$0.40&89.68$\pm$0.31  & 92.87$\pm$0.27 &95.56$\pm$0.12  &93.21$\pm$0.18 & 79.36$\pm$0.58 &45.50$\pm$0.12 \\
        GraphMAE & 84.20$\pm$0.40 & 73.40$\pm$0.40 &  81.10$\pm$0.40&90.02$\pm$0.24  & 93.19$\pm$0.39 & 95.53$\pm$0.14 &92.25$\pm$0.23 & 80.52$\pm$0.32 & 49.35$\pm$0.54 \\
        SeeGera & 83.93$\pm$0.23 & 73.0$\pm$0.80 &  80.10$\pm$0.42&88.39$\pm$0.26  & 92.81$\pm$0.45 & 95.39$\pm$0.08 &93.84$\pm$0.11 & 79.23$\pm$0.45 & 49.20$\pm$0.54 \\
        NWR-GAE & 83.62$\pm$1.61 & 71.45$\pm$2.41 & \underline{83.44$\pm$0.92} &\underline{90.12$\pm$0.15}  & \underline{93.53$\pm$0.18} & 95.55$\pm$0.12&93.23$\pm$0.34&79.83$\pm$0.26 & 47.39$\pm$0.56\\
        GREET &83.82$\pm$0.48  &73.22$\pm$0.64  & 80.20$\pm$0.75 &88.02$\pm$0.58 & 92.45$\pm$0.77 &95.43$\pm$0.34&\underline{93.85$\pm$0.33} &\underline{80.68$\pm$0.31} & 48.64$\pm$0.30\\
        GraphACL & 84.20$\pm$0.31 & \underline{73.63$\pm$0.22} & 82.02$\pm$0.15 & 89.80$\pm$0.25 & 93.31$\pm$0.19 & 95.35$\pm$0.16&92.11$\pm$0.33&79.66$\pm$0.29& \underline{50.12$\pm$0.45} \\
         AUG-MAE & 84.30$\pm$0.40 & 73.20$\pm$0.42 & 81.12$\pm$0.38 & 90.08$\pm$0.12 & 93.34$\pm$0.52 & 95.55$\pm$0.17 &92.56$\pm$0.20 &80.53$\pm$0.34& 50.26$\pm$ 0.35 \\
         Bandana & \underline{84.62$\pm$0.37} & 73.60$\pm$0.16 & \textbf{83.53$\pm$0.51} & 89.62$\pm$0.09 & 93.44$\pm$0.11 & \underline{95.57$\pm$0.04} &93.10$\pm$0.05 &80.58$\pm$0.44& 49.80$\pm$ 0.23 \\
        \textbf{DGMAE} & \textbf{84.93$\pm$0.51} & \textbf{73.82$\pm$0.64} & 81.10$\pm$0.42 & \textbf{90.75$\pm$0.47} & \textbf{93.96$\pm$0.41} & \textbf{96.01$\pm$0.12} & \textbf{93.93$\pm$0.20} & \textbf{81.46$\pm$0.21} & \textbf{51.56$\pm$0.33}\\        
        \bottomrule
    \end{tabular}}
       \label{table.2}
\end{table*}

\subsection{Overall Loss}
The sub-loss $\mathcal{L}_f$ learns the contextual representation for each mask node by predicting the original features, and the sub-loss $\mathcal{L}_d$ captures the discrepancies from the context for each mask node. The overall loss of our model is a combination of the two sub-losses:
\begin{equation}
    \mathcal{L} = (1-\lambda) \cdot \mathcal{L}_f + \lambda \cdot \mathcal{L}_d,
\label{eq.13}
\end{equation}
where $\lambda$ is a balance hyperparameter, the two modules can progressively enhance each other during training. In each training epoch, the node set is divided into two parts, the decoder extracts the common information to recover the mask features by the visible nodes. In this process, the mutual information of the mask node with its neighbors is maximized. The feature reconstruction optimizes the decoder so that it is equipped to extract the node-neighborhood common information. The unmasked node set extracts the unique representation through discrepancy reconstruction. Therefore, these two branches can be co-trained well together. See the Appendix \ref{loss analysis} for more analysis of the loss function.

\begin{table*}[ht]
    \centering
     \caption{Node clustering performance on the heterophilic graph. The clustering performance is evaluated by four metrics with mean value and standard deviation. }
   \begin{tabular}{c|c|ccccccc}
        \toprule
          Dataset & Metrics &Dink-Net & NWR-GAE & MaskGAE & GraphMAE&GraphACL & GREET&\textbf{DGMAE} \\
        \midrule
        \multirow{4}{*}{Texas}
        &ACC & 41.97±3.67 &\underline{55.03$\pm$4.43} &46.78$\pm$4.73&46.28$\pm$2.41 &46.50$\pm$2.75 & 52.40$\pm$3.77&\textbf{66.83$\pm$6.19}\\
        &NMI & 7.27±2.04 & 12.97$\pm$2.70& 7.66$\pm$2.08&6.06$\pm$2.14 &7.58$\pm$1.26&\underline{37.19$\pm$4.95}&\textbf{37.41$\pm$7.20}\\
        & ARI & 5.80±2.68 & 17.73$\pm$5.31 &11.27$\pm$5.44& 0.99$\pm$3.29  & 11.12$\pm$1.39&\underline{24.50$\pm$5.11} & \textbf{42.49$\pm$13.39}  \\
        &F1 &  27.11±3.86 & 29.83$\pm$2.85 & 27.95$\pm$2.57 & 21.48$\pm$2.64 & 27.54$\pm$1.68&\underline{41.79$\pm$6.67} & \textbf{42.19$\pm$4.04} \\
        \midrule
        \multirow{4}{*}{Cornell}
        &ACC& 35.03±1.55 &36.12$\pm$2.35 & 40.98$\pm$2.42&46.12$\pm$2.91 &38.31$\pm$4.22&\underline{48.96$\pm$3.90}&\textbf{57.16$\pm$5.99}\\
        &NMI & 11.61±2.14 &3.36$\pm$0.54 & 12.91$\pm$2.23&12.07$\pm$3.25 &7.71$\pm$1.78&\underline{28.19$\pm$3.34}&\textbf{34.41$\pm$7.88}\\
        & ARI& 4.70±2.67 & 0.51$\pm$1.20 & 6.30$\pm$2.15& 8.66$\pm$3.01  &2.05$\pm$2.27&\underline{23.00$\pm$5.26}& \textbf{29.66$\pm$9.92} \\
        &F1 & 29.05±2.14 & 20.41$\pm$1.14 & 29.28$\pm$3.56& 28.24$\pm$4.03 & 25.54$\pm$2.52&\underline{42.43$\pm$4.13} & \textbf{43.22$\pm$5.13}  \\
        \midrule        
        \multirow{4}{*}{Wisconsin}
        &ACC & 45.66±2.54 &48.53$\pm$1.80 & 42.55$\pm$3.75&52.67$\pm$4.64 &41.31$\pm$5.05&\underline{56.65$\pm$5.62} &\textbf{68.69$\pm$4.52}\\
        &NMI & 15.79±2.47 &9.35$\pm$1.40 & 10.03$\pm$2.19&18.54$\pm$3.11 &9.31$\pm$2.04& \underline{22.57$\pm$6.20} &\textbf{44.19$\pm$2.19}\\
        & ARI & 12.58±3.35 & 11.57$\pm$2.35 & 4.52$\pm$2.23 & 16.66$\pm$2.96  & 5.11$\pm$3.51&\underline{21.36$\pm$6.50} & \textbf{40.56$\pm$4.13} \\
        &F1 & 30.86±2.92 & 25.46$\pm$3.32 & 27.49$\pm$2.20& 34.52$\pm$3.12 &29.31$\pm$3.45&\underline{36.83$\pm$7.29} &\textbf{48.81$\pm$4.17}  \\
        \midrule
        \multirow{4}{*}{Chameleon}
        &ACC & 27.35±0.41 & 27.01$\pm$0.64 & 30.70$\pm$1.78&\underline{37.02$\pm$1.50} &30.98$\pm$1.91&34.44$\pm$1.93 &\textbf{39.96$\pm$1.74} \\
        &NMI& 5.42±0.24 & 3.65$\pm$0.88 & 5.96$\pm$1.52&\underline{19.47$\pm$1.53} &6.73$\pm$1.37&14.99$\pm$2.61 &\textbf{22.39$\pm$0.43} \\
        & ARI &2.56±0.18 & 1.63$\pm$0.48 & 3.44$\pm$0.84&\underline{14.39$\pm$1.79} &  4.38$\pm$1.86&10.59$\pm$1.59 & \textbf{17.05$\pm$0.79} \\
        &F1 &23.49±0.35 & 20.53$\pm$1.34 &28.09$\pm$2.19&\underline{28.61$\pm$2.53} &24.14$\pm$02.45&27.24$\pm$2.86  & \textbf{29.61$\pm$2.89} \\
        \midrule
        \multirow{4}{*}{Squirrel}
         &ACC & 23.83±1.17 & 23.81$\pm$0.34 & 26.19$\pm$1.25&\underline{27.61$\pm$0.33} &24.45$\pm$0.95&25.64$\pm$1.99 &\textbf{30.83$\pm$0.71} \\
        &NMI & 1.28±0.86 & 0.97$\pm$0.18 & 2.79$\pm$1.06&\underline{4.58$\pm$0.33} &1.66$\pm$0.81&3.22$\pm$0.97&\textbf{8.00$\pm$0.73} \\
        & ARI  &0.65±0.77 & 0.52$\pm$0.12 & 2.20$\pm$0.95&\underline{3.63$\pm$0.26} & 0.93$\pm$0.64&1.32$\pm$0.49& \textbf{6.40$\pm$0.66} \\
        &F1  &22.04±1.35 & 20.38$\pm$0.59 &\underline{23.27$\pm$1.99}&22.80$\pm$1.24& 17.94$\pm$1.77&15.61$\pm$3.84 & \textbf{26.12$\pm$2.18} \\
        \midrule
        \multirow{4}{*}{Roman}
        &ACC& 16.50±0.34 & 18.82$\pm$0.91 &15.76$\pm$0.34&15.38$\pm$0.36&15.36$\pm$0.34&\underline{31.43$\pm$0.78}&\textbf{36.25$\pm$0.86}\\
        &NMI &8.49±0.12 &12.81$\pm$0.37 & 12.90$\pm$0.31&9.33$\pm$0.20&12.08$\pm$0.38&\underline{37.00$\pm$0.74}&\textbf{37.57$\pm$0.53}\\
        & ARI &4.56±0.14 & 5.78$\pm$0.45 & 4.14$\pm$0.30& 3.64$\pm$0.13 & 4.24$\pm$0.20&\underline{14.05$\pm$0.74}& \textbf{23.31$\pm$0.86} \\
        &F1 & 12.32±0.32 &  15.03$\pm$0.81 & 12.01$\pm$0.38&13.60$\pm$0.41 & 11.64$\pm$0.30&\underline{24.00$\pm$0.98} & \textbf{30.08$\pm$0.91} \\      
        \bottomrule
    \end{tabular}
    \label{table.cluster}
\end{table*}

\section{Experiment}

\subsection{Datasets}
For node classification, we performed experiments on homophilic and heterophilic graphs. For homophilic graphs, we employed Cora, Citeseer, Pubmed~\cite{gcn, collective}, WikiCS~\cite{wiki},  Amazon Computer, Amazon Photo, CoAuthor CS, CoAuthor Physics~\cite{cs, photo}, and Flicker~\cite{graphsaint-iclr20}. For heterophilic graphs, we employed Texas, Cornell, Wisconsin~\cite{geom-gcn}, Chameleon, Squirrel, Crocodile~\cite{squchame}, and Roman Empire~\cite{roman}. We also provide node classification for larger heterophilous graphs in Appendix \ref{newheter}. Details of the datasets are in the Appendix \ref{Datasets}. For graph classification, the results are shown in the Appendix \ref{grapgclasification}.

\subsection{Baselines}
We consider two categories of baselines as follows: (1) Contrastive learning methods: DGI~\cite{dgi}, CCA-SSG~\cite{ccassg}, BGRL~\cite{BGRL}, GCA~\cite{GCA}, Infograph~\cite{infograph}, SP-GCL, GREET~\cite{greet}, GraphACL~\cite{graphacl}; (2) Generative learning methods: VGAE~\cite{vgae}, SeeGera~\cite{seegera}, GraphMAE~\cite{graphmae}, MaskGAE~\cite{maskgae}, DSSL~\cite{dssl}, NWR-GAE~\cite{NWR-GAE}, AUG-MAE~\cite{augmae}, Bandana~\cite{bandana}, S2GAE~\cite{s2gae}, GraphMAE2~\cite{graphmae2}. Refer to Appendix \ref{baselines} for a description of baselines.


\subsection{Experimental Settings}
\label{Experiment set}
For node classification, we follow the settings of the commonly used linear evaluation methods e.g.,~\cite{dgi}. First, the model is trained in an unsupervised manner. Then, we freeze the node representations and train a logistic regression classifier, We report accuracy on the node classification task. For node clustering, we adopt four evaluation metrics, including ACC, NMI, ARI, and F1, which are widely used in previous methods~\cite{cluster,DCRN,liuyue_deep_graph_clustering_survey,RGC}. We randomly conduct each experiment on each dataset 10 times and calculate the mean and standard deviation. For the graph classification task, we follow the same settings in previous works~\cite{s2gae,graphmae,augmae}. We run experiments on a machine with an NVIDIA TITAN RTX GPU with 24GB of GPU memory. Detailed hyperparameter settings and their analysis are in Appendix \ref{hyperparameter} and Appendix \ref{hyper_experiment}.

\subsection{Node Classification Performance Comparison}
The node classification results for 8 heterophilic and 9 homophilic graph datasets are reported in Table \ref{table.1} and Table \ref{table.2}, respectively. DGMAE performs best in 16 out of 17 benchmarks on all self-supervised learning methods. Compared to previous generative methods that performed poorly on heterophilic graphs, our method made significant improvements, especially on heterophilic graphs, while our method achieved comparable performance on homophilic graphs. GraphMAE with feature reconstruction as a goal is ineffective on heterophilic graphs. Since feature reconstruction essentially minimizes the similarity between nodes and their neighbors, this results in high similarity of node pairs. AUG-MAE and Bandana employ advanced masking strategies, but the performance improvement for node classification of the heterophilic graph is limited. In contrast, DGMAE can prevent the indistinguishable representations between heterophilic pairs of nodes by learning feature discrepancies between nodes, maintaining distinguishability between different classes of nodes.

\subsection{Node Clustering Performance Comparsion}
We conduct node clustering on the heterophilic graph to evaluate the quality of the learned representations.  It's more challenging than clustering on the homophilic graph.  Specifically, we perform 10 k-means clustering for each dataset and calculate the mean and variance. The experimental results are reported in Table \ref{table.cluster}.  Compared with the state-of-the-art self-supervised learning methods, our method achieves promising performance on four metrics since the discrepancy reconstruction helps the model capture the discrepancy of nodes. Especially on the three datasets texas,cornell, and wisconsin, the performance of DGMAE is much higher than the second-best method GREET. Meanwhile, we find that the NMI and ARI metrics of our clustering results significantly improve compared to the sub-optimal performance, which indicates that the clustering results have greater mutual information and consistency with the real labels. The corresponding results also demonstrate the superiority of our proposed DGMAE and the effectiveness in various downstream tasks.

\begin{table}[t]
     \caption{The effect of the different components.}
        \begin{tabular}{l|c c c c}
        \toprule
        Baseline  & Roman & Actor & Citeseer &Photo \\
        \midrule
       variant 1 & 75.35$\pm$0.30&33.41$\pm$1.07& 73.03$\pm$0.52&93.40$\pm$0.42   \\
        variant 2 &36.71$\pm$0.50&28.14$\pm$1.12&73.05$\pm$0.23&93.15$\pm$0.41\\
        variant 3 &75.04$\pm$0.44&35.16$\pm$0.75&58.02$\pm$1.33&92.88$\pm$0.45 \\  
        DGMAE &\textbf{76.62$\pm$0.41}&\textbf{36.61$\pm$0.74}&\textbf{73.82$\pm$0.64}&\textbf{93.96$\pm$0.41} \\  
        \bottomrule
    \end{tabular}
        \label{components}
\end{table}

\subsection{Ablation Study}
To examine the effectiveness of each component and its contribution to DGMAE, we conducted ablation experiments on several variants of the model, (1) variant 1: removing the edge selection module(w/o discrepancy selection), (2)variant 2: removing the feature reconstruction(w/o feature reconstruct), (3)variant 3: removing the discrepancy reconstruction(w/o discrepancy reconstruct). Table \ref{components} shows that discrepancy reconstruction is highly effective on heterophilic graphs, while feature reconstruction is mainly successful on homophilic graphs. This supports our argument that feature reconstruction depends on relevant contextual common information. In heterophilic graphs, most neighborhoods belong to different classes, lacking sufficient common information between nodes to recover mask features, which may hinder the reconstruction of the original features. On the other hand, discrepancy reconstruction potentially reduces the similarity between similar nodes in homophilic graphs by capturing node disparities. The discrepancy selection module, which precedes discrepancy reconstruction, focuses on extracting features of dissimilar node pairs by utilizing heterophilic edges to reduce the influence of homophilic edges. The results indicate that incorporating both heterophilic and homophilic information, which is inevitable in any graph, can enhance overall performance.

\subsection{Sensitivity Analysis of Different Encoders}
We migrated GNNs specifically designed for heterophilous graphs as encoders ~\cite{h2gnn,aero-gnn} to the existing GMAE to verify the efficiency of the discrepancy reconstruction, and despite the improvement over the original, the performance on heterophilous graphs was not as good as the superior performance achieved in supervised learning. For a more in-depth study, we migrated these encoders to DGMAE, and the results show that the addition of discrepancy reconstruction loss results in improvement on all datasets. However, it does not reflect an advantage over the classical GNN encoder, which suggests that the optimization of the pretext task of self-supervised learning also plays an important role. Among the existing GMAEs, GraphMAE based on feature reconstruction and MaskGAE based on link reconstruction have been proved to implicitly align connected node representations~\cite{augmae}, leading to node representations becoming similar. Therefore, we think that while replacing the state-of-the-art encoder in self-supervised learning can improve the performance of GMAE on heterophilous graphs, designing a pretext task that is compatible with both homophilous and heterophilous graphs will be more generalizable.

\begin{table}[h]
\renewcommand{\arraystretch}{1.2} 
\setlength{\tabcolsep}{1.5pt} 
\small 
\centering
\caption{The performance on different encoders.}
\begin{tabular}{l|cccc}
\toprule
Method & Texas & Cornell & Wisconsin  & Roman\\
\midrule
GMAE           & $55.14\pm6.75$     & $51.62\pm6.45$     & $55.69\pm3.06$     & $46.73\pm0.55$     \\
GMAE (GCN)    & $52.35\pm7.28$     & $59.19\pm5.98$     & $61.57\pm5.56$     & $48.80\pm0.64$     \\
GMAE (H2GCN)    & $66.76\pm5.14$     & $51.89\pm7.53$     & $74.90\pm4.79$     & $58.81\pm0.50$     \\
GMAE (AERO-GNN) & $62.70 \pm 5.38$     & $51.62 \pm 5.33$     & $62.75 \pm 5.33$     & $59.24 \pm 0.39$     \\
\midrule
DGMAE              & $\textbf{88.11$\pm$5.16}$ & $\textbf{78.65$\pm$4.59}$ & $\textbf{88.43$\pm$3.56}$ & $\textbf{76.62$\pm$0.41}$ \\
DGMAE (GCN)       & $83.53\pm5.26$     & $68.38\pm3.64$     & $84.31\pm4.56$     & $64.76\pm0.53$     \\
DGMAE (H2GCN)       & $76.49\pm3.43$     & $58.65\pm7.36$     & $77.84 \pm 5.19$     & $75.68\pm0.47$     \\
DGMAE (AERO-GNN)    & $80.00 \pm 5.82$     & $65.41 \pm 7.03$     & $77.31 \pm 4.25$     & $68.84 \pm 0.45$     \\
\midrule
H2GCN              & $84.86\pm6.77$    & $82.16\pm4.80$    & $86.67\pm4.69$    & $60.11\pm0.52$     \\
AERO-GNN           & $84.35 \pm 5.20$      & $81.24 \pm 6.80$      & $84.80 \pm 3.30$      & $72.46 \pm 0.37$     \\
\bottomrule
\end{tabular}
\label{tab:my_label}
\end{table}

\begin{figure}[h]
  \subfigure[Cora]{
      \includegraphics[width=0.45\columnwidth]{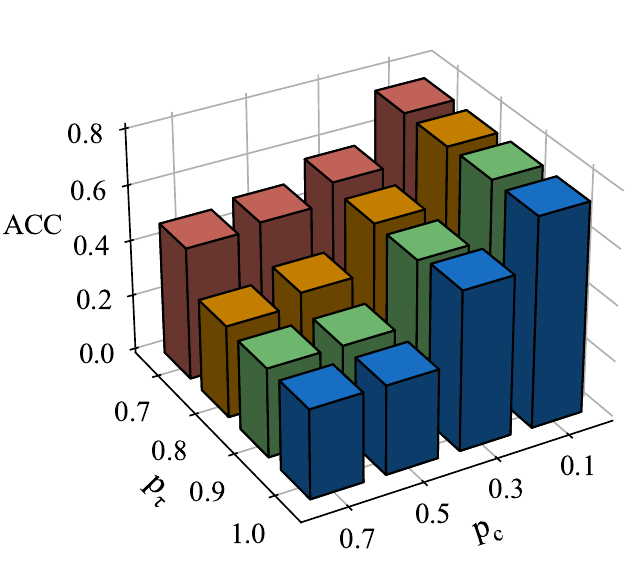}}
  \subfigure[Roman]{
      \includegraphics[width=0.45\columnwidth]{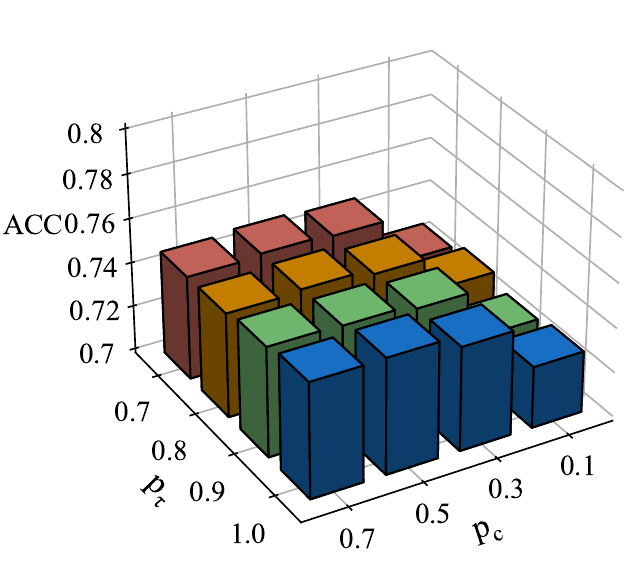}}
    \caption{The effect of $p_c$ and $p_\tau$ on homophilic graph and heterophilic graph}
  \label{pc_pt}
   \Description{..}
\end{figure}

\subsection{Sensitivity Analysis of Removing Ratio \texorpdfstring{$p_c$}{p_c} and Cut-Off Ratio $p_\tau$}
As shown in Figure \ref{pc_pt}, the effects of $p_\tau$ and $p_c$ on homophilic and heterophilic graphs are investigated, with the range of the $p_\tau$ threshold set to [0.7, 0.8, 0.9, 1.0] and the range of the $p_c$ set to [0.1, 0.3, 0.5, 0.7]. $p_t$ can prevent the destruction of semantic information caused by excessive $p_c$. Additionally, $p_c$ exhibits the opposite tendency on homophilic and heterophilic graphs. It is important to note that the proportion of heterophilic edges is much smaller than homophilic graphs. The role of $p_c$ is to screen out edges with large feature discrepancies in pairs of nodes. However, setting $p_c$ too high may interfere with the discrepancy reconstruction of nodes in the same class. In heterophilic graphs, increasing $p_c$ can provide more comprehensive information on feature discrepancies, as most node pairs belong to different classes. This emphasizes the significance of heterophilic edges in the calculation of feature discrepancies.

\begin{figure}[h]
 \setlength{\abovecaptionskip}{0.cm}
  \centering
  \subfigure[Roman]{\includegraphics[width=0.32\columnwidth]{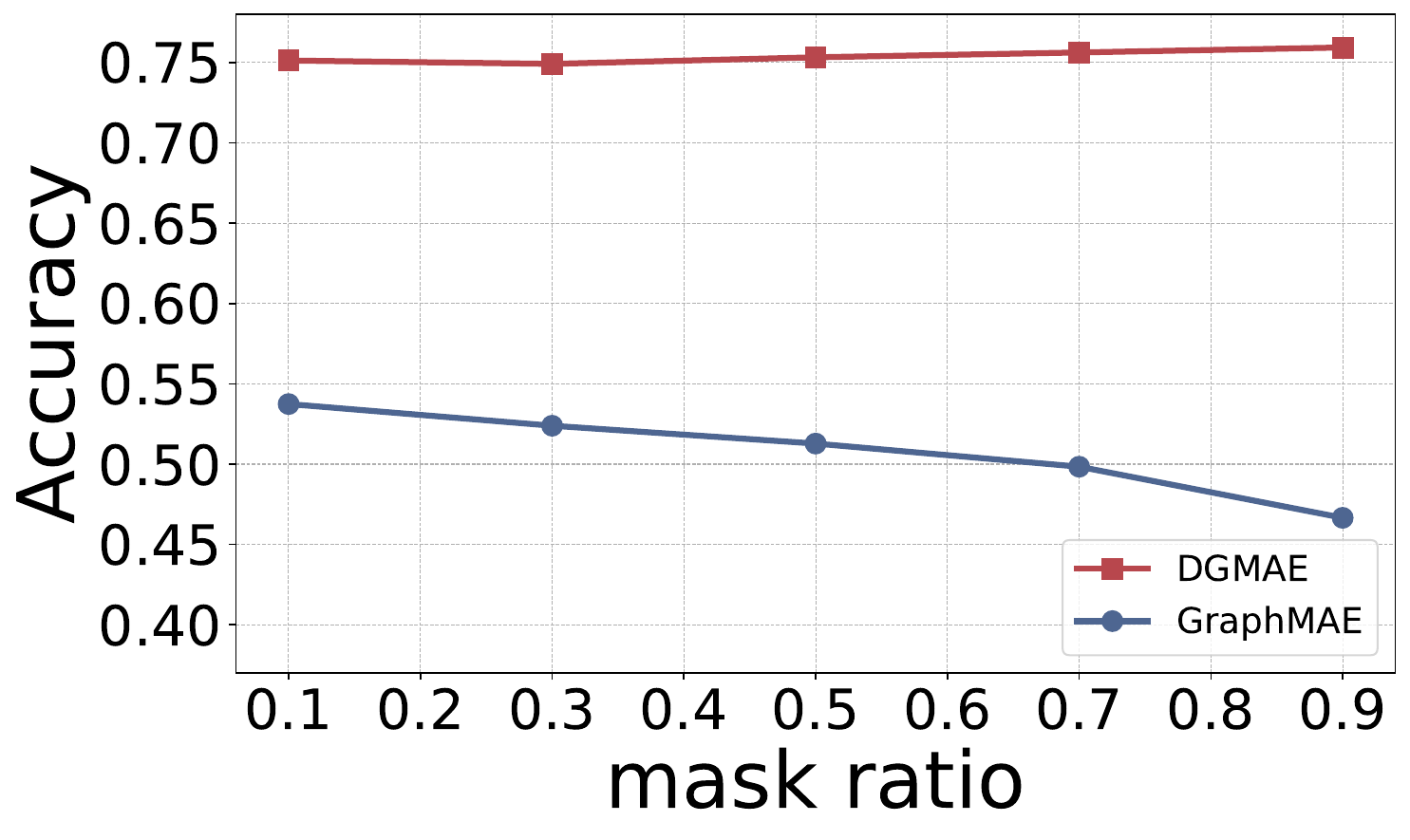}}
  \subfigure[Squirrel]{\includegraphics[width=0.32\columnwidth]{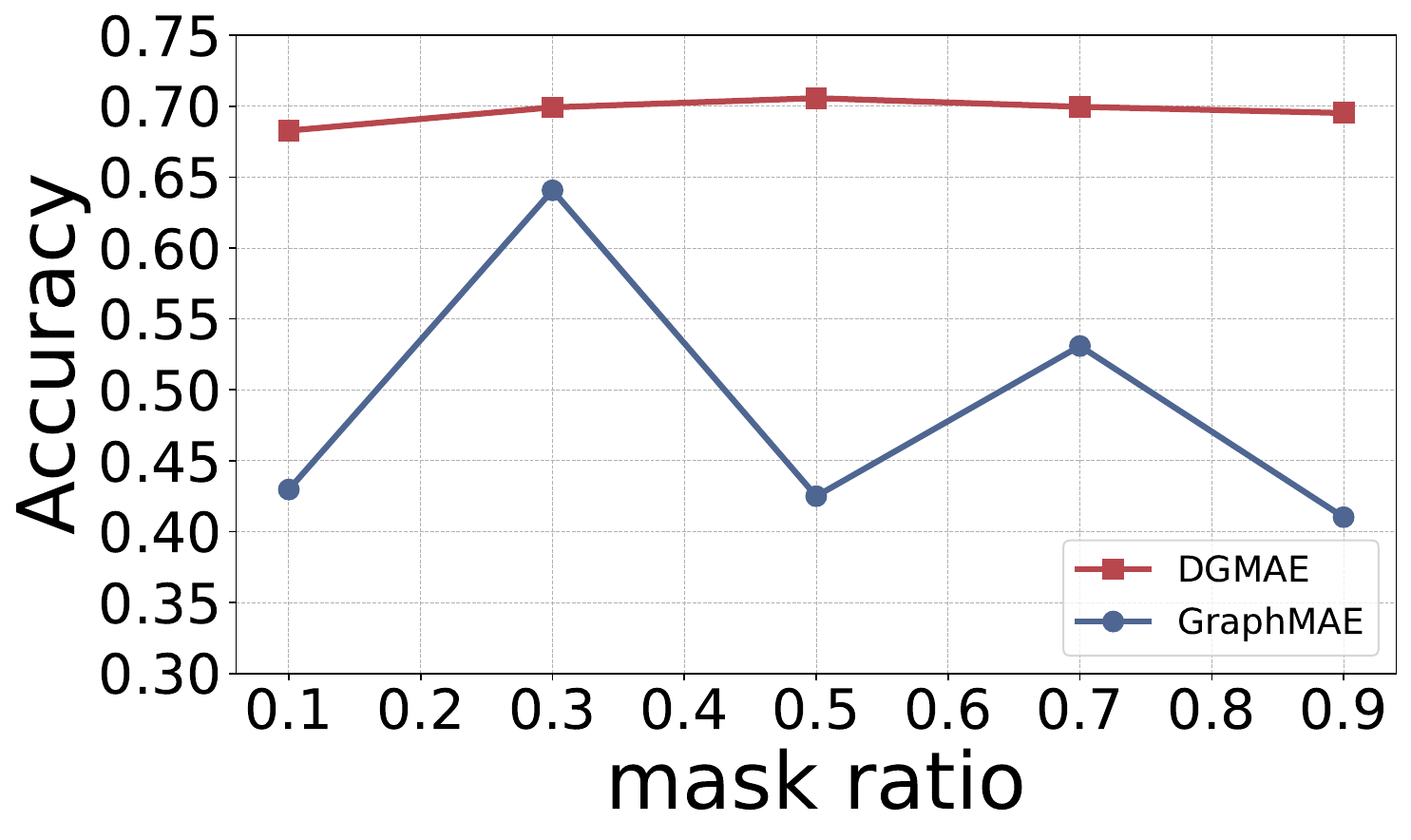}}
    \subfigure[Chameleon]{\includegraphics[width=0.32\columnwidth]{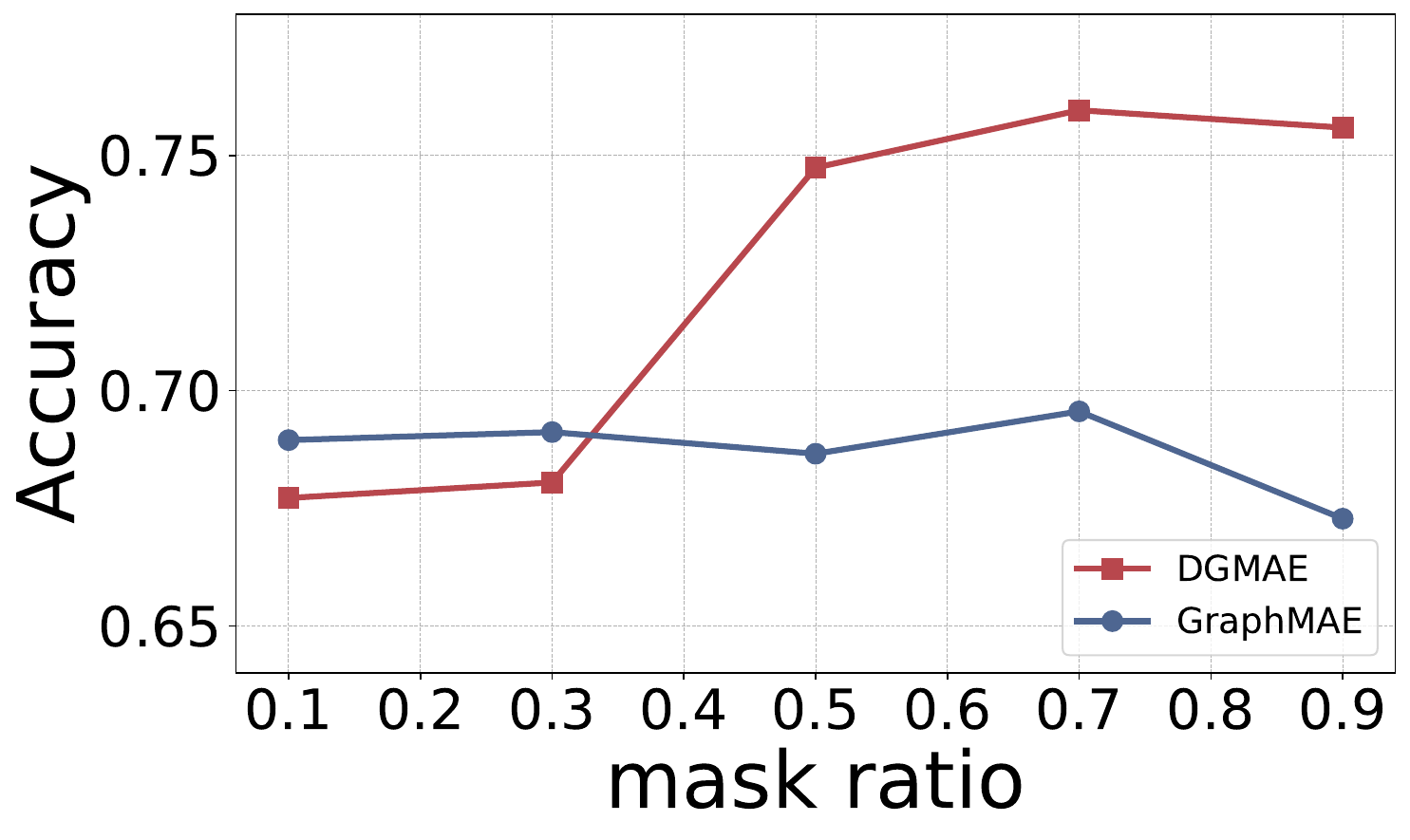}}
  \caption{The effect on the heterophilic graph of different mask ratios.}
  \label{maskr_atio}
     \Description{..}
\end{figure}

\subsection{Effect of Mask Ratio}
To further investigate the effect of the mask ratio on the discrepancy reconstruction, we retain only the discrepancy loss and study the model's performance under different mask ratios. Observing the curve changes of GraphMAE and DGMAE under different heterophilic graph datasets in Figure \ref{maskr_atio}, we find that in GraphMAE, the model performance decreases as the mask rate increases. This is because an excessively high mask ratio results in masked nodes with less contextual information available for reconstruction. In contrast, our approach still has superior performance at a high mask ratio. In heterophilic graphs (e.g., Roman, Squirrel), DGMAE maintains excellent performance (Roman: $75.48\% \rightarrow73.5\%$) even with a masking rate as high as 80\%, while the traditional method GraphMAE drops significantly ($46.73\%\rightarrow38.2\%$). This suggests that the discrepancy reconstruction mechanism is able to capture key discrepancy information from sparse contexts rather than relying on dense local neighbor information. Although it lacks relevant information about the reconstructed nodes themselves, high mask ratios imply that the neighborhood discrepancy information becomes larger, which is the ability to learn distinguishable node representations by capturing the discrepancies between nodes through discrepancy reconstruction. It shows that DGMAE can learn expressive node representations in visible information scarcity.

\subsection{Effect of Adaptive Discrepancy Selection}
To demonstrate the effectiveness of the adaptive discrepancy selection component in discrepancy reconstruction, we fix $p_c=0.2,p_\tau=1.0$ and use only the discrepancy reconstruction loss. The sampling method is set as (1) original attention weight $w_{ij}$; (2) reversed attention weights (1-$w_{ij}$); (3) random sampling; and (4) degree centrality of the edge score. The experiments are conducted on the three heterophilous graph datasets. The experimental results show that adaptive discrepancy sampling based on reversed attention weights is the best, original attention weight sampling is the worst, and random sampling lies in between. This demonstrates the importance and effectiveness of sampling high feature discrepancy edges, whereas sampling based on degree centrality only takes into account structural importance and does not consider feature discrepancy. This experiment illustrates that adaptive discrepancy selection plays a positive role in discrepancy reconstruction.
\begin{table}[h]
\caption{The performance of differnet sampling method.}
\centering
\begin{tabular}{l|ccc}
 \toprule
sampling method & Actor & Chameleon & Squirrel \\
\midrule
$w_{ij}$ & 34.34±0.41 & 70.88±1.09 & 58.20±1.98 \\
$1-w_{ij}$ & \textbf{35.36±0.68} & \textbf{73.90±1.92} & \textbf{65.61±1.35} \\
random sample & 34.72±0.74 & 71.80±2.13 & 60.07±1.67 \\
degree sample & 35.12±0.64 & 73.44±1.53 & 60.63±1.16 \\
\bottomrule
\end{tabular}
\label{tab:adaptive discrepancy}
\end{table}
\subsection{Effect of Discrepancy on Representation Uniformity}
\label{uniformity}
\begin{figure}[ht]
 \setlength{\abovecaptionskip}{0.cm}
  \centering
  \subfigure[$\lambda=0.0$]{\includegraphics[width=0.24\columnwidth]{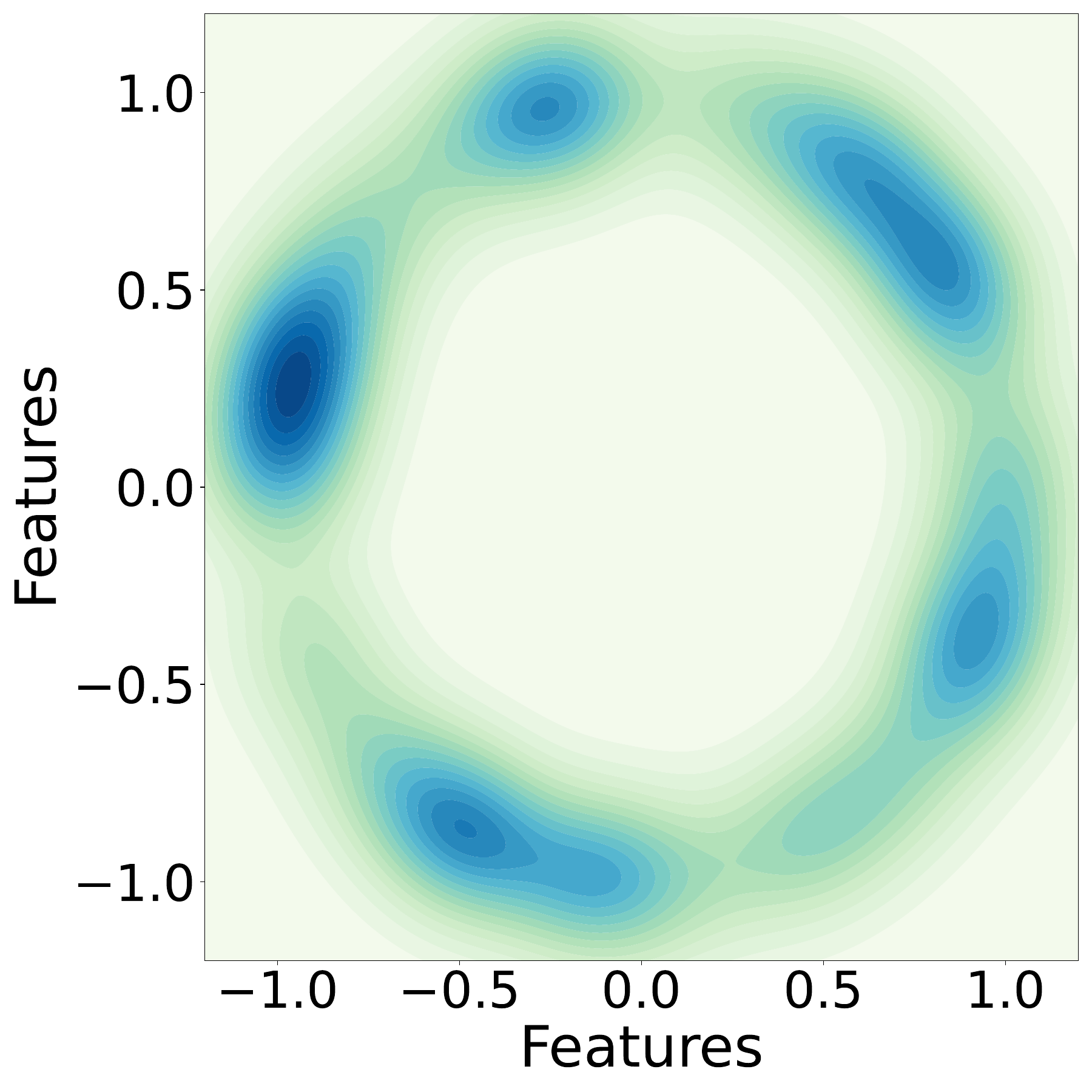}}
  \subfigure[$\lambda=0.1$]{\includegraphics[width=0.24\columnwidth]{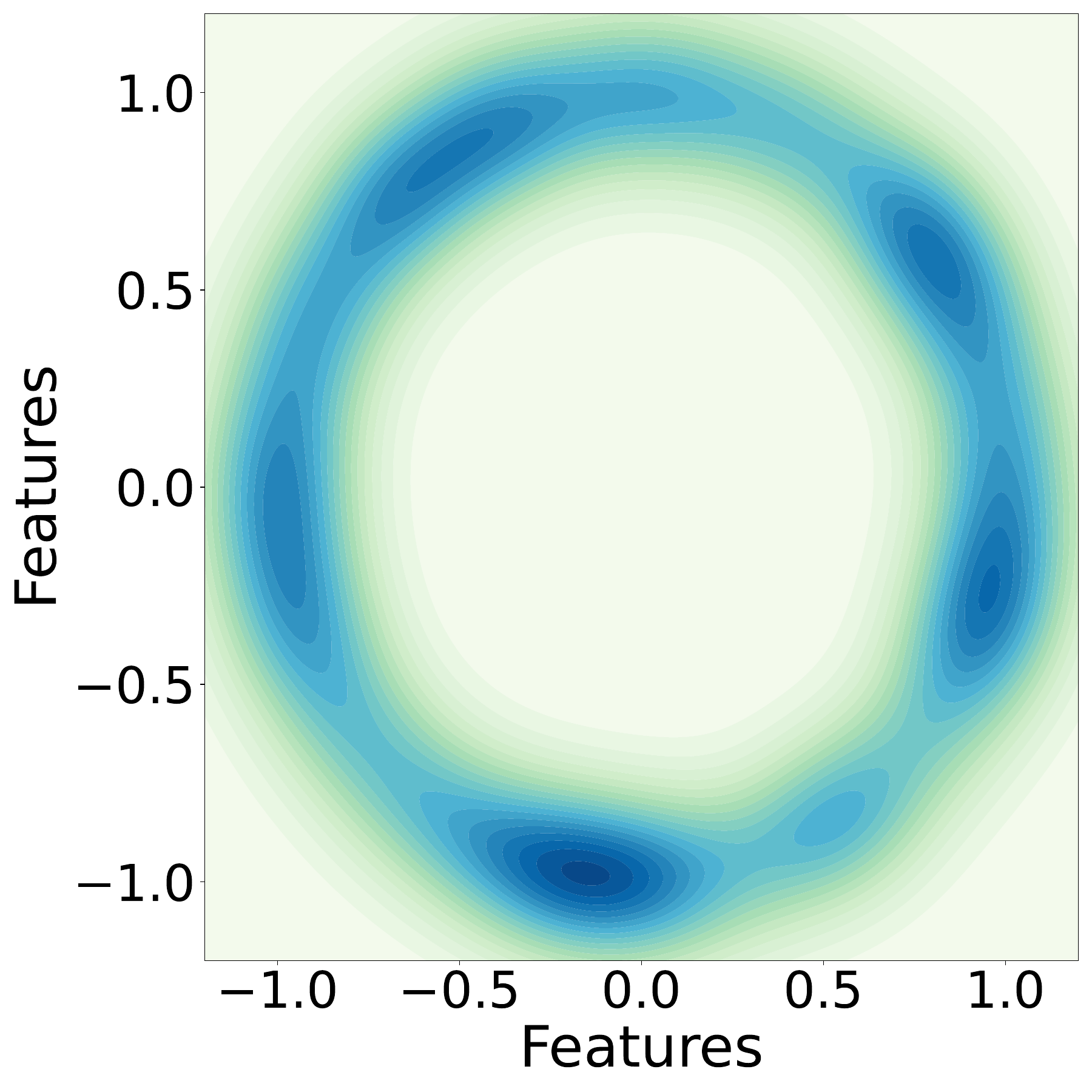}}
    \subfigure[$\lambda=0.5$]{\includegraphics[width=0.24\columnwidth]{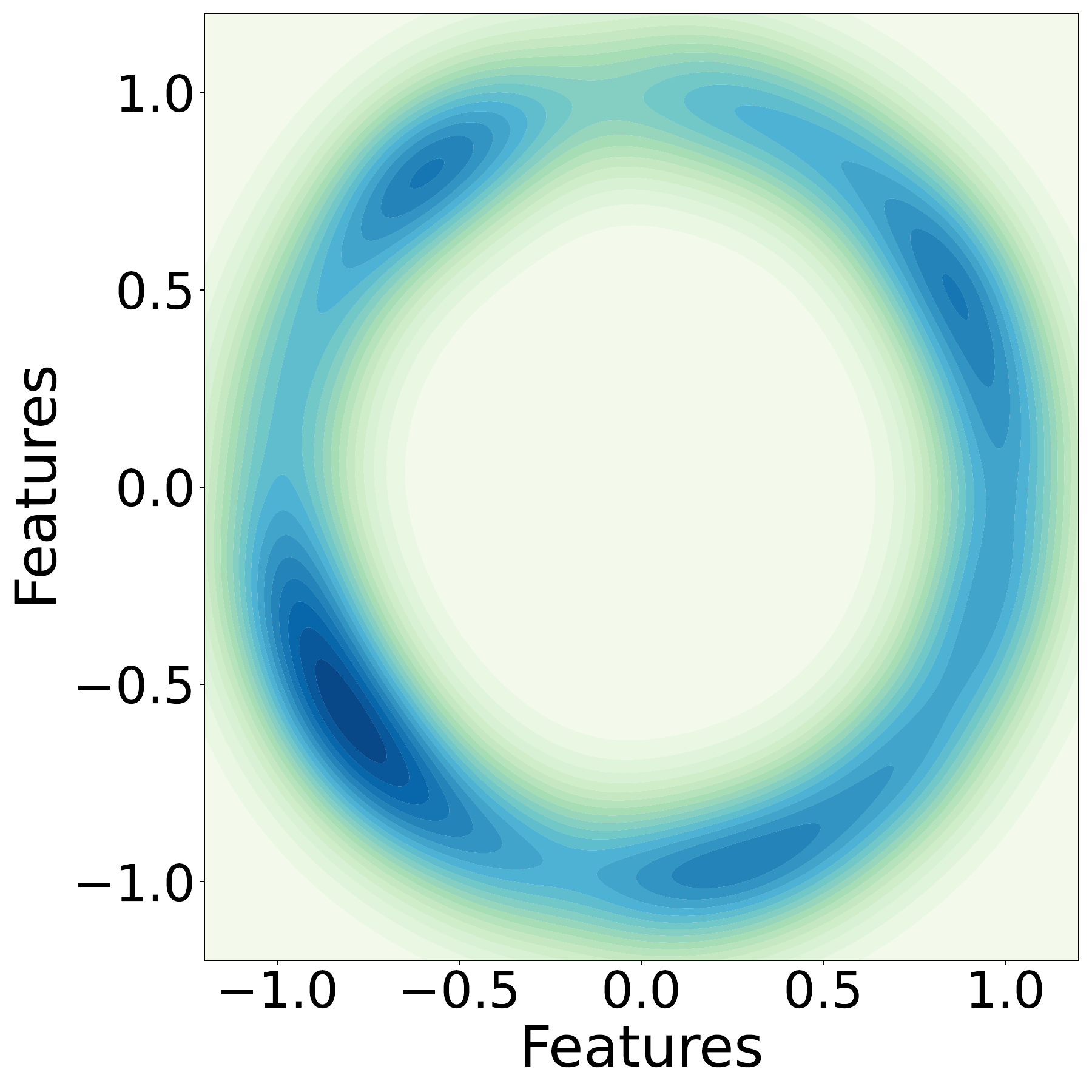}}
    \subfigure[$\lambda=1.0$]{\includegraphics[width=0.24\columnwidth]{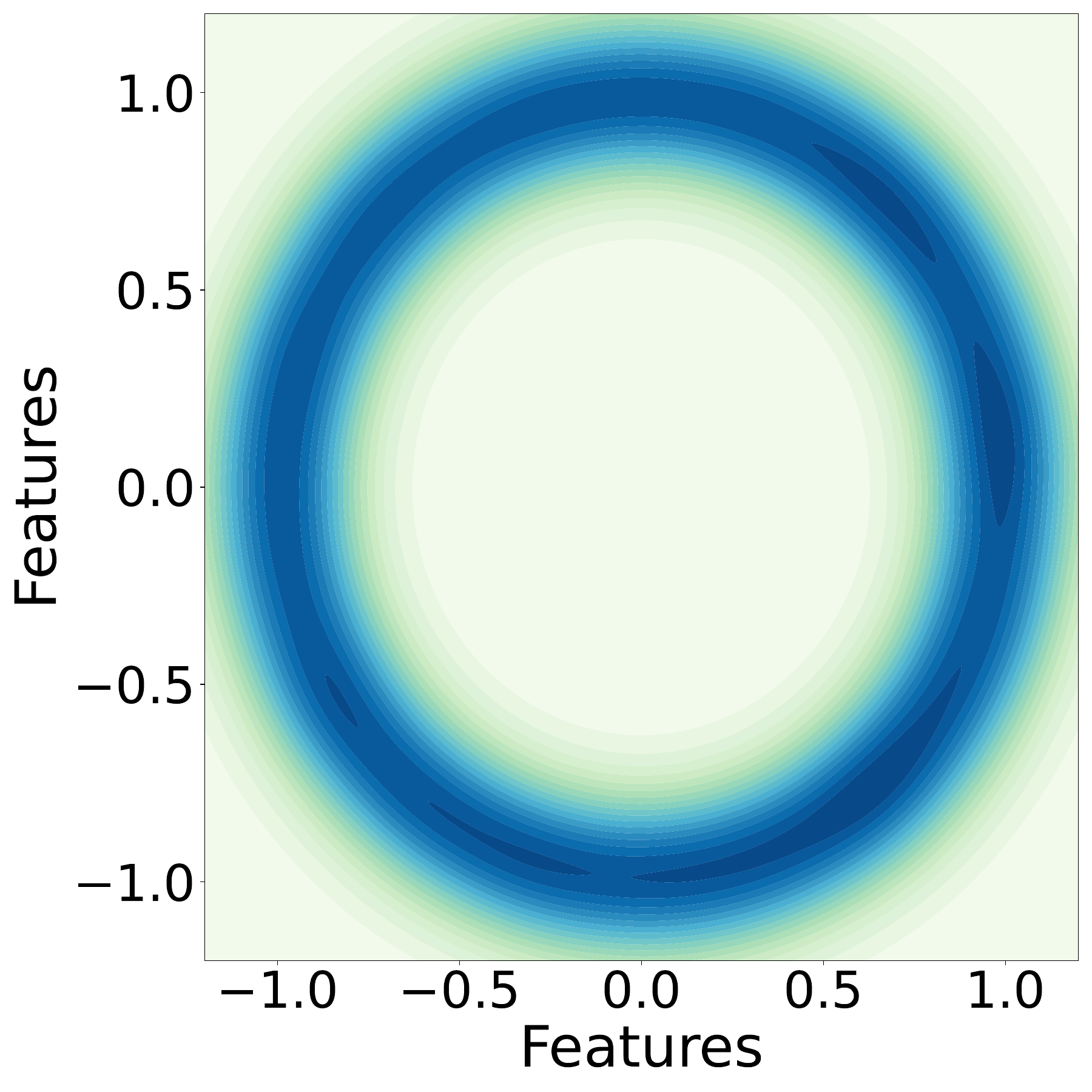}}
  \caption{Representation distributions of Cora on S1 learned
 by different discrepancy weights. We plot feature distributions with Gaussian kernel density estimation (KDE) in $\mathcal{R}^2$.}
  \label{Uniformity1}
     \Description{..}
\end{figure}

\begin{figure}[ht]
 \setlength{\abovecaptionskip}{0.cm}
  \centering
  \subfigure[$\gamma_2=1$]{\includegraphics[width=0.24\columnwidth]{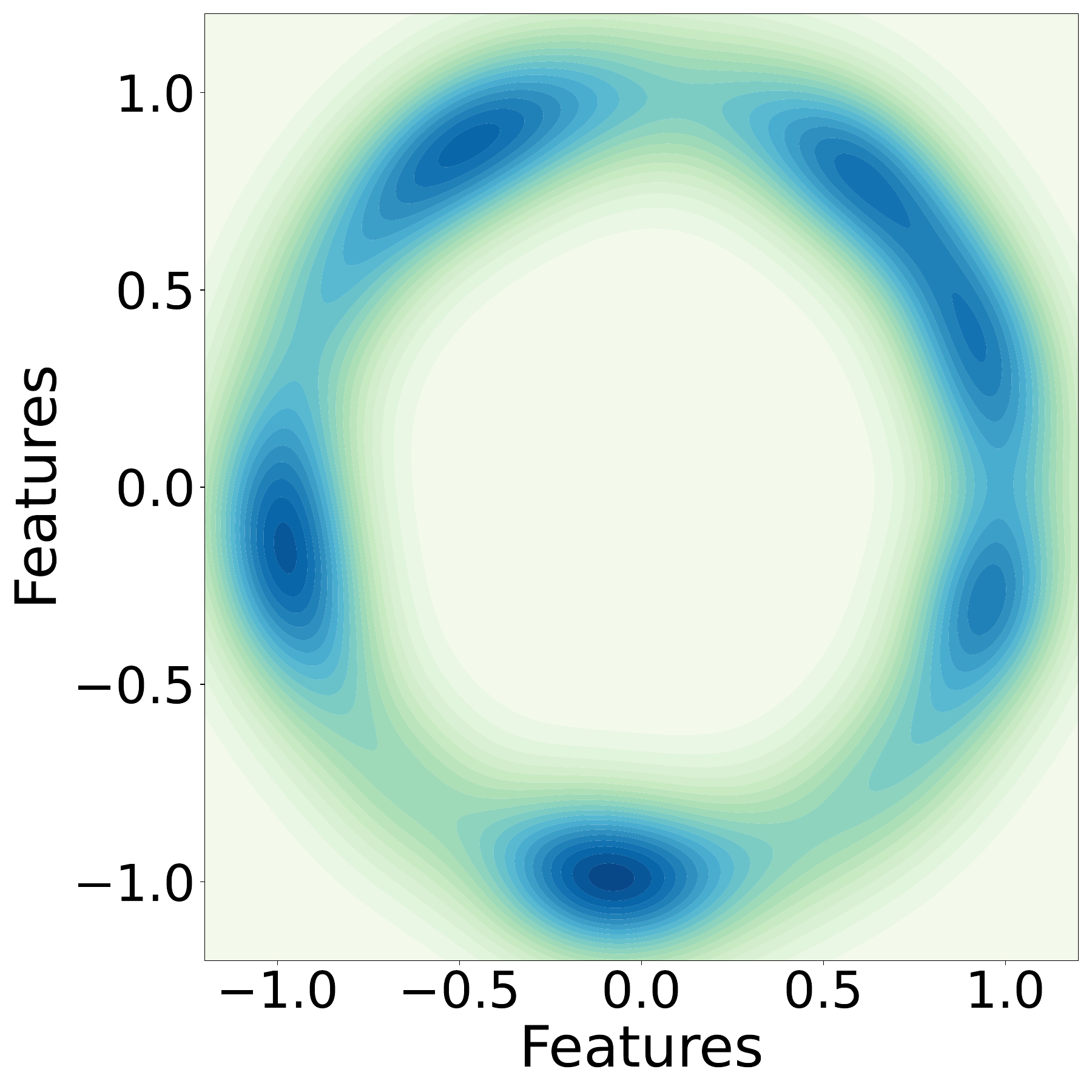}}
  \subfigure[$\gamma_2=5$]{\includegraphics[width=0.24\columnwidth]{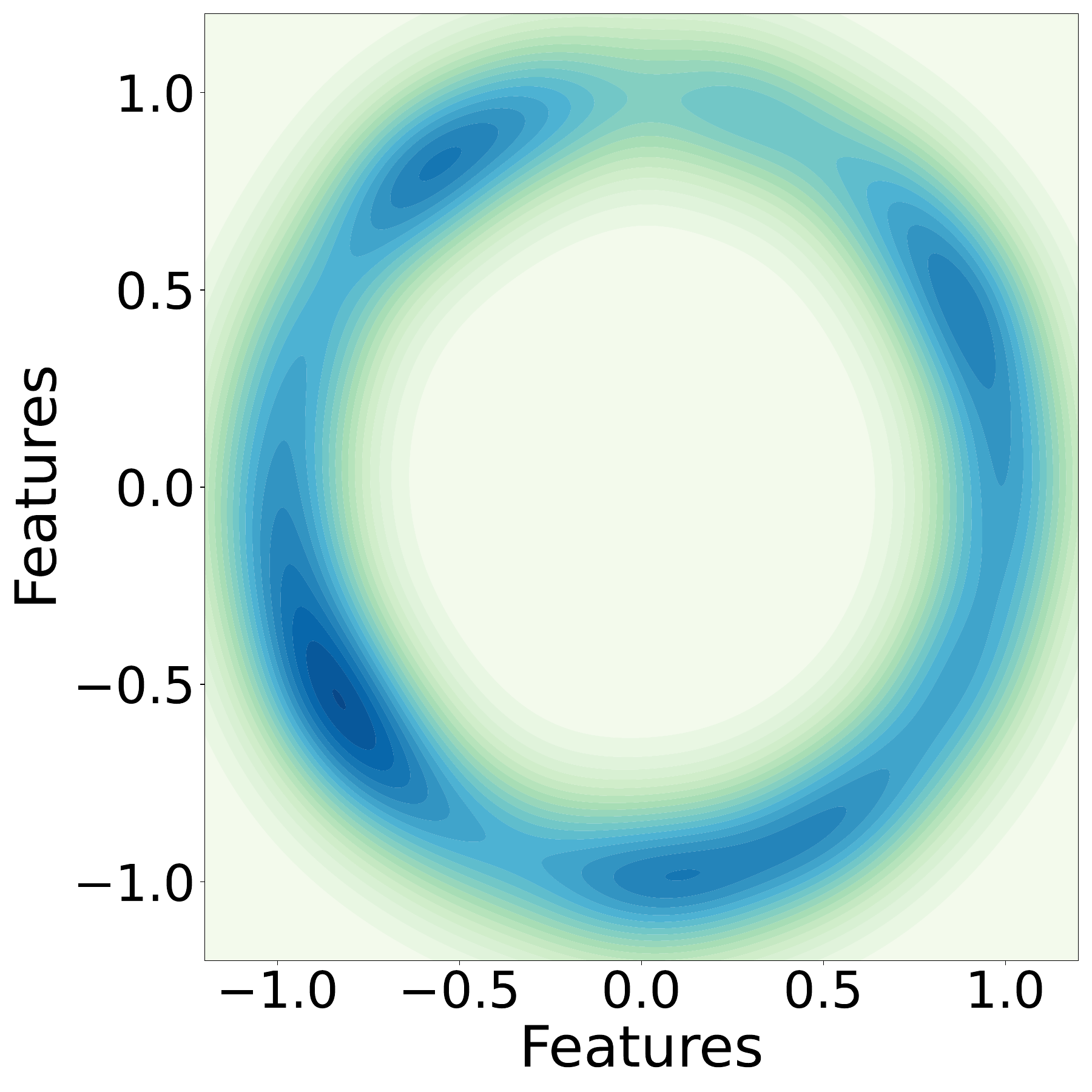}}
    \subfigure[$\gamma_2=10$]{\includegraphics[width=0.24\columnwidth]{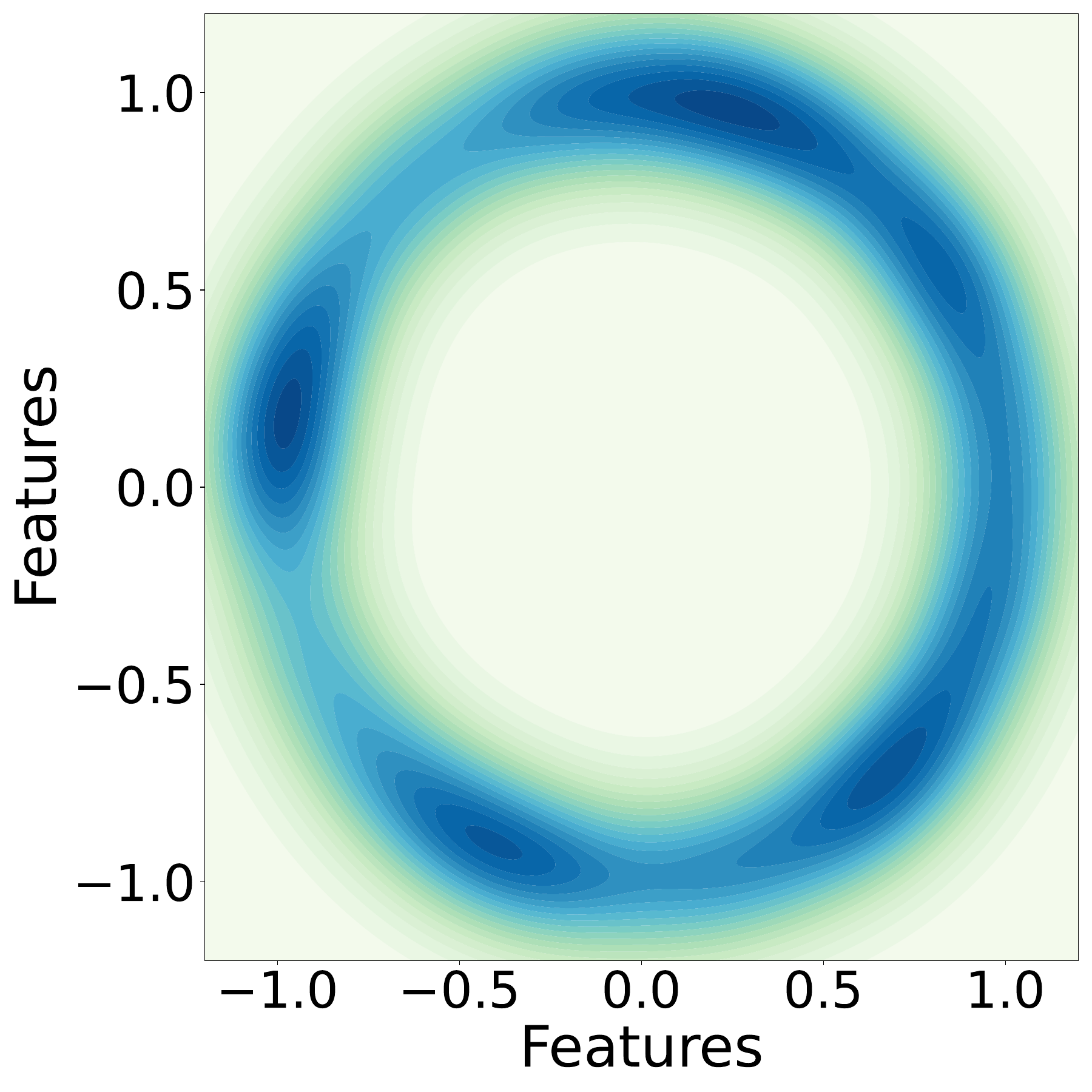}}
    \subfigure[$\gamma_2=15$]{\includegraphics[width=0.24\columnwidth]{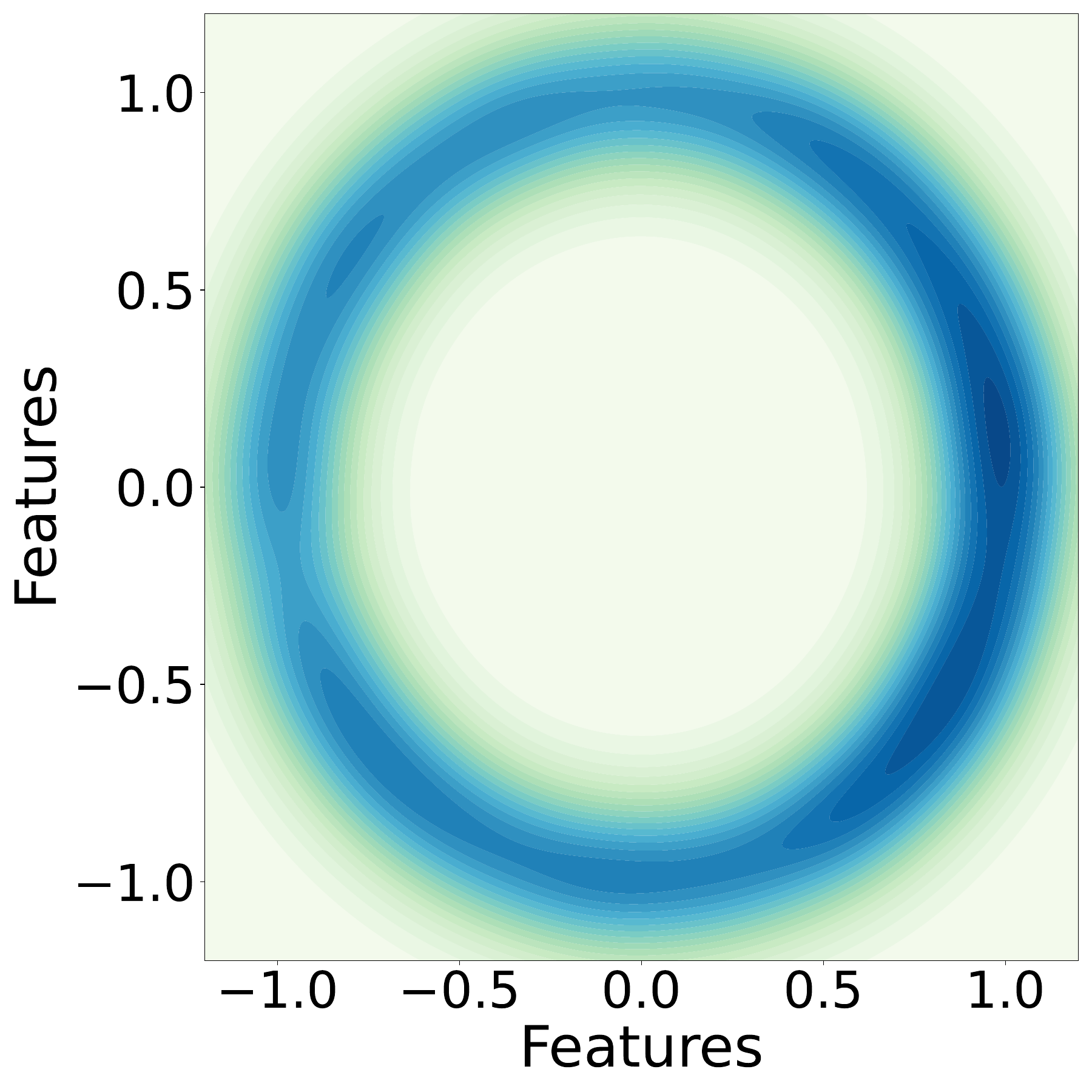}}
  \caption{Representation distributions of Cora on S1 learned
 by different values of $\gamma_2$. We plot feature distributions with Gaussian kernel density estimation (KDE) in $\mathcal{R}^2$.}
  \label{Uniformity2}
     \Description{..}
\end{figure}

We introduce $\lambda$ in Eq.13 to balance feature reconstruction and discrepancy losses. In Eq.12, we add a scaling coefficient $\gamma_2$ for the discrepancy loss to adjust node weights based on reconstruction error. We study their effects on representation distribution uniformity during discrepancy reconstruction by visualizing at different values. As shown in Figure \ref{Uniformity1}, the distribution of representations obtained is more uniform as the weight of discrepancy weights increases, which indicates that different nodes are distributed further and further apart on the hypersphere. It is consistent with our conclusion in Section \ref{pairrep} that the discrepancy loss can push the distance between samples farther away.
As shown in Figure \ref{Uniformity2}, The scaling coefficients that regulate disparity loss also affect the uniformity of representations, with representations with higher scaling coefficients focusing more on learning subtle discrepancies and obtaining a more uniform distribution of representations. The above shows that the difference loss improves the representation uniformity by increasing the distance between different samples, thus improving the quality of the node representation.

\vspace{-0.5cm}

\subsection{Visualization of Pairwise Representation}
\label{pairrep}
\begin{figure}[ht]
 \setlength{\abovecaptionskip}{0.cm}
  \centering
  \subfigure[Roman]{\includegraphics[width=0.325\columnwidth,height=0.254\columnwidth]{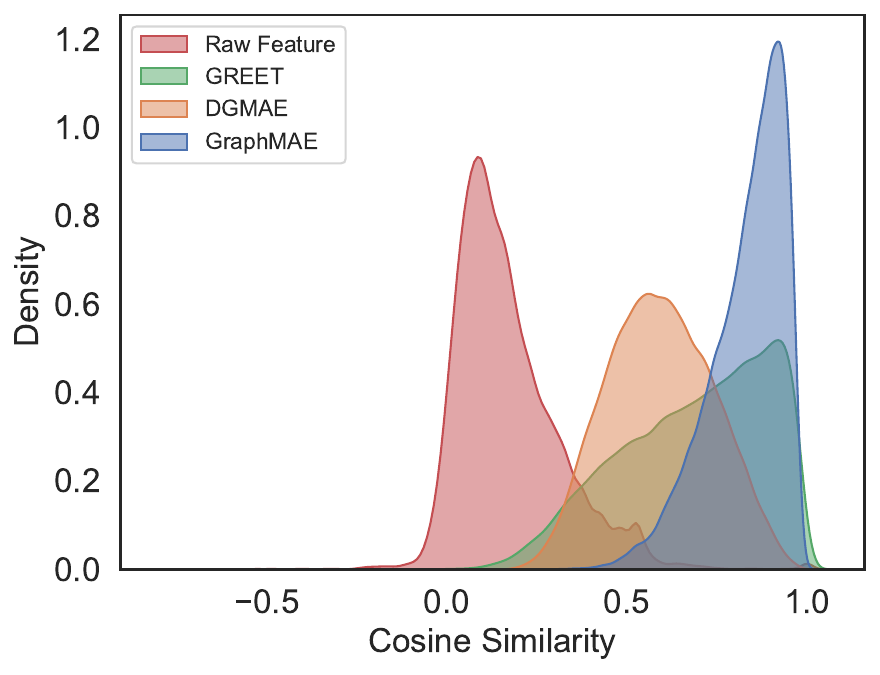}}
  \subfigure[Squirrel]{\includegraphics[width=0.325\columnwidth,height=0.25\columnwidth]{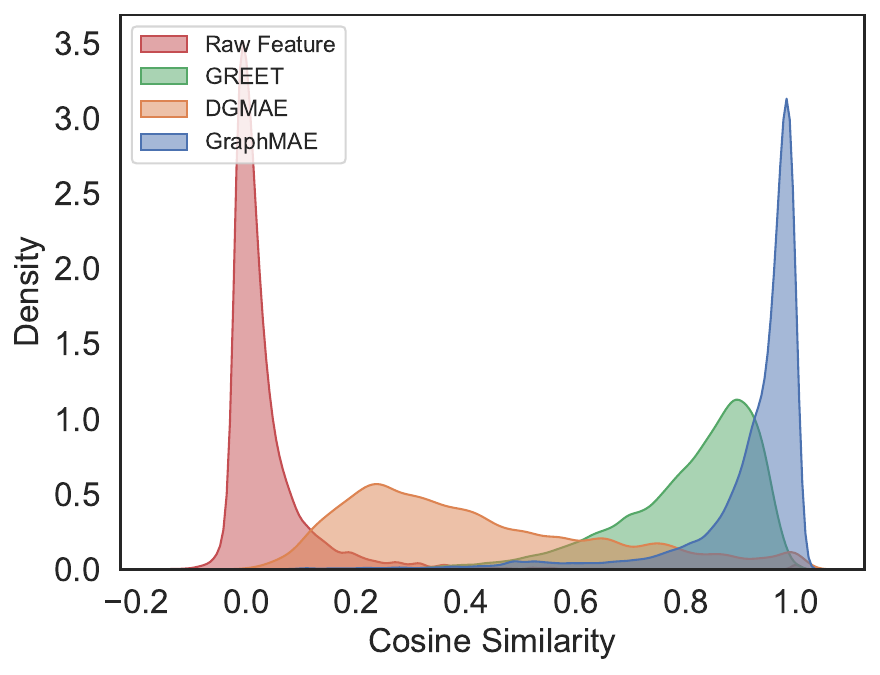}}
  \subfigure[Actor]{\includegraphics[width=0.325\columnwidth,height=0.25\columnwidth]{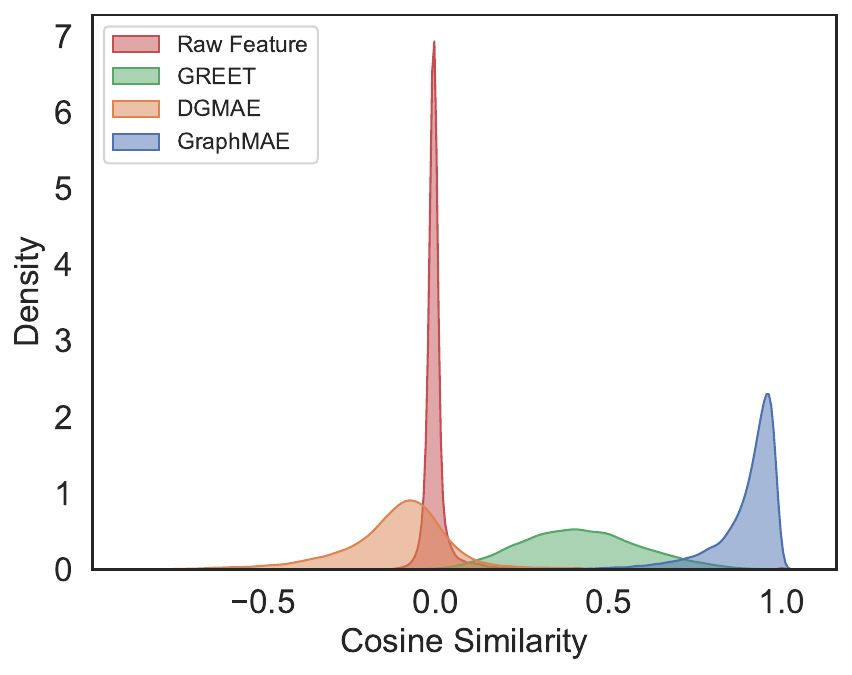}}
  \caption{Similarity distribution of heterophilic node pairs.}
  \label{fig.heteropair}
       \Description{..}
\end{figure}

To investigate the property of the node representations learned by DGMAE, we visualize the pairwise representation similarity of Roman, Squirrel, and Actor compared to GraphMAE. As shown in Figure \ref{fig.heteropair},  GraphMAE enforces similar representations between nodes and their neighbors through feature reconstruction, resulting in higher similarity between pairs of nodes in a heterophilic graph. Compared with GraphMAE, it can be observed that the distribution of pairwise similarity learned by DGMAE shifts to the left. Our approach is more clearly distinguishable between these heterophilic node pairs compared with GREET, indicating that it can effectively suppress the similarity of node pairs to prevent the features of different classes of nodes from converging.

\section{Conclusion}
In this work, we note that existing GMAE methods lack awareness of the discrepancy information between nodes, affecting their effectiveness on heterophilic graphs. To this end, we propose a novel masked graph generative learning method named DGMAE. The key is to capture the nodes' discrepancy between nodes to avoid node confusion. Specifically, we define a learnable discrepancy feature as the reconstruction target which preserves the original discrepancy in the low-dimensional space. The model reconstructs not only the original features but also the discrepancy features. Experiments show that DGMAE significantly outperforms state-of-the-art methods on heterophilic graphs and is adaptable to multiple downstream tasks, demonstrating superior effectiveness and generalizability.

\begin{acks}
This work was supported in part by the National Natural Science Foundation of China under Grants 62303366, 62133012, 62425605, 62472340, and 62203354, in part by the Key Research and Development Program of Shaanxi under Grants 2025CY-YBXM-041, 2024GX-YBXM-122, and 2022ZDLGY01-10, and in part by the Fundamental Research Funds for the Central Universities under Grants ZYTS25086, ZYTS25211, and ZYTS25090.
\end{acks}

\bibliographystyle{ACM-Reference-Format}
\balance
\bibliography{DiffGAE}

\clearpage
\appendix
\section{Datasets}
\label{Datasets}
We introduce the details of the datasets and baselines as follows. and the statistics of them are shown in Table \ref{dataset}.

\begin{table*}[ht]
    \centering
        \caption{Statistics of datasets.}
    \begin{tabular}{c|cccccc}
        \toprule
        Dataset  & Node & Edges & Classes &  Features & Train/Val/Test & Edge homophily $h_{edge}$ \\
        \midrule
        Cora &  2078&5278&7&1433&140/500/1000& 0.81\\
        Citeseer &  3327&4552&6&3703&120/500/1000&0.74\\
        Pubmed &   19717& 44324&3& 500&60/500/1000& 0.80\\
        Computer  &   13752 &574418& 10 &767 &10\%/10\%/80\% & 0.78\\
        Photo &  7650&119081&8&745&10\%/10\%/80\% &0.83\\
        CS & 18333&163788&15&6805&10\%/10\%/80\%& 0.81\\
        Physics &  34493&495924&5&8415&10\%/10\%/80\%& 0.93\\
        WikiCS &  11701&431206&10&300&10\%/10\%/80\% & 0.65\\
        Flicker &  89250&899756&7&500&50\%/25\%/25\% & 0.32\\
        Texas &  183&309&5&1703&48\%/32\%/20\% &0.11\\
        Cornell &  183&295&5&1703&48\%/32\%/20\% & 0.30\\
        Wisconsin &  251&466&5&1703&48\%/32\%/20\% & 0.21\\
        Chameleon  &  2277&36101&5&2325&48\%/32\%/20\% &0.23\\
        Crocodile &  11631&360040&5&2325&60\%/20\%/20\% & 0.25\\
        Squirrel &  5201&216933&5&2089&48\%/32\%/20\% & 0.22\\
        Actor &  7600&33544&5&931&48\%/32\%/20\% & 0.22\\
       Roman &32927&5278&18&300&48\%/32\%/20\% &0.05\\
        \bottomrule
    \end{tabular}
    \label{dataset}
\end{table*}

\textbf{Cora, Citeseer, and Pubmed}~\cite{citation,gcn}. Three classical citation networks, where each node in the graph represents a paper, the edges represent the citation relations of the paper, the features are bag-of-words representations of the paper, and the node labels are the research areas of the paper. All three are highly homophilic graphs and are widely used in node classification.

\textbf{Computer and Photo}~\cite{photo}. The graph dataset based on Amazon.com products, where nodes represent products, edges represent co-purchase relationships between products, node features are product reviews, and labels correspond to different products, is used to evaluate the model's classification performance for different product categories.

\textbf{CS and Physics}~\cite{cs}. The graph dataset is based on the Academic Collaboration Network, derived from the Microsoft Academic Graph. The nodes represent authors and the edges indicate collaborative relationships between two authors. The node features consist of bag-of-words representations of the keywords of their papers, and node labels indicate the research areas.

\textbf{WikiCS}~\cite{wiki}. It is derived from data from Wikipedia articles in the field of computer science. The nodes represent articles, while the edges represent hyperlinks between articles. The features are averaged from the text of the corresponding articles using pre-trained GloVe~\cite{glove} word embeddings, and the labels represent different branches of the computer science field.

\textbf{Flicker}~\cite{graphsaint-iclr20}. The dataset is from NUS-wide, where nodes denote images uploaded to Flickr, and edges denote two images with some common attributes. The node features are bag-of-words representations provided by NUS, and the labels are the categories of the images. The task of this dataset is to classify image types based on descriptions and common attributes of online images.

\textbf{Texas, Cornell, and Wisconsin}~\cite{geom-gcn}. They are three heterophilic networks collected by Carnegie Mellon University from computer science departments at different universities, where nodes are computer science web pages and edges are hyperlinks between them. The node characteristics are bag-of-words representations of the web pages, and the labels indicate the types of the web pages.

\textbf{Chameleon, Squirrel, and Crocodile}~\cite{squchame}. They are heterophilic networks based on the Wikipedia network, where nodes denote pages in Wikipedia and edges denote links between them. The nodes features are informative nouns on Wikipedia pages and the labels denote the average traffic of the pages.

\textbf{Actor}~\cite{geom-gcn}. It's an actor co-occurrence network where nodes denote actors and edges denote that two actors have co-occurrences in the same film. The nodes features denote keywords on the Wikipedia page, and on the actor's Wikipedia page, the tags are classified into five classes.

\textbf{Roman-empire (Roman)}~\cite{roman}. The dataset is based on a strongly heterophilic network of Roman Empire articles from the English Wikipedia, where each node corresponds to a (non-unique) word in the text, and edges represent the context and dependency tree relationships of the word. The node features are the word embeddings of the words. The node labels are the syntactic roles corresponding to each word.

\section{Baselines}
\label{baselines}
\textbf{DGI}~\cite{dgi}: It is an unsupervised learning model for graph neural networks that learns high-quality representations of nodes by maximizing the mutual information of node representations and graph summary.

\textbf{GCA}~\cite{GCA}: It is a graph contrastive learning method that introduces an adaptive enhancement to improve performance from feature and topology considerations.

\textbf{CCA-SSG}~\cite{ccassg}: It is a contrastive learning method to reduce different view correlations by introducing typical correlation analysis

\textbf{BGRL}~\cite{BGRL}: It is a graph contrastive learning method using two different encoders to learn node representations via graph augmentation and momentum updating without introducing negative samples.

\textbf{SP-GCL}~\cite{sp-gcl}: It is a single-pass graph contrastive learning method with the single-pass augmentation-free graph contrastive learning loss.

\textbf{Greet}~\cite{greet}: It is a graph contrastive learning method that is compatible with homography and heterography to learn two different views of high pass signals and low pass signals through an edge discriminator.

\textbf{GraphACL}~\cite{graphacl}: It is a graph contrastive learning method capturing one-hop local neighborhood information and two-hop similarity by learning neighborhood distributions.

\textbf{VGAE}~\cite{vgae}: It is a graph variational self-encoder that uses variational inference to learn a low-dimensional representation of graph data.

\textbf{GraphMAE}~\cite{graphmae}: This is a graph generative learning method, which learns a low-dimensional representation of a graph by masking node features and reconstructing these masked portions.

\textbf{MaskGAE}~\cite{maskgae}: This is a graph generative learning method that proposes masked graph modeling as a pretext task to reduce information redundancy through path masking strategies.

\textbf{DSSL}~\cite{dssl}: This is a self-supervised learning method that learns the neighborhood distribution of a node by decoupling the different underlying semantics between different neighborhoods.

\textbf{NWR-GAE}~\cite{NWR-GAE}: This is a graph generative learning method to reconstruct the whole neighborhood information about proximity and structure through Neighbourhood Wasserstein Reconstruction.

\textbf{Infograph}~\cite{infograph}: This is a self-supervised learning method based on graph-level representations, which learns unsupervised graph representations by maximizing the mutual information between graph-level representations and substructure representations at different scales.

\textbf{GrahCL}~\cite{graphacl}: This is a graph comparison learning method that learns node embeddings by maximizing the similarity between the representations of two randomly perturbed versions of a local subgraph of the same node.

\textbf{S2GAE}~\cite{s2gae}: This is a graph generative learning method that learns generalizable node representations by reconstructing masked edges through directed edge masks and cross-correlation decoders

\textbf{GraphMAE2}~\cite{graphmae2}: This is a graph generative learning method that applies regularisation to feature reconstruction to improve the discriminability of input features, through a strategy of multi-view random remask decoding and latent representation prediction to normalize feature reconstruction.

\textbf{AUG-MAE}~\cite{augmae}: This is a graph generative learning method that proposes an adaptive masking strategy from simple to difficult to learn to align difficult positive samples. Uniformity regularisation is also introduced to ensure the uniformity of the learned representation.

\textbf{Bandana}~\cite{bandana}: This is a graph generative learning method that proposes a non-discrete edge masking strategy: bandwidth masking instead of binary masking and better access to topology information by predicting bandwidths.

\section{Hyperparameter Settings}
\label{hyperparameter}
The detailed hyperparameter settings are listed in Table~\ref{hyper}.
\begin{table}[ht]
\renewcommand\arraystretch{1.2}
    \centering
        \caption{Details of the hyperparameters.}
      \resizebox{\linewidth}{!}{\begin{tabular}{c|cccccccccc}
        \toprule
        Dataset  & $\lambda$&$p_c$ & $p_\tau$ & $\gamma_1 $& $\gamma_2$ &weight\_delay&learning\_ rate & mask\_ratio & num\_layer \\
        \midrule
        Cora &  0.1&0.3&0.7&3&6&2e-4& 1e-4&0.5&2\\
        Citeseer &  0.1&0.3&0.7&3&4&5e-7&5e-5&0.5&2\\
        Pubmed &   0.1&0.1&0.9&3&1&1e-5 & 1e-3&0.75&2\\
        Computer   &  0.1&0.1&0.9&3&3&2e-4& 1e-3&0.5&2\\
        Photo  &  0.1&0.3&0.7&3&5&2e-4& 1e-3&0.5&2\\
        CS &  0.4&0.1&0.9&3&1&5e-5& 1e-3&0.7&2\\
        Physics  &  0.4&0.1&0.9&3&1&5e-5& 1e-3&0.5&2\\
        WikiCS  &  0.4&0.1&0.9&3&1&1e-3& 1e-4&0.5&2\\
        Flicker  &  0.9&0.3&0.7&3&3&2e-4& 1e-3&0.5&2\\
        Texas  &  0.8&0.3&0.7&3&3&2e-4& 1e-4&0.75&1\\
        Cornell &  0.8&0.5&0.6&3&3&2e-4& 1e-4&0.2&1\\
        Wisconsin  & 0.4&0.3&0.7&3&5&5e-4& 1e-4&0.75&1\\
        Chameleon   &  0.5&0.3&0.7&3&3&2e-4& 1e-4&0.75&2\\
        Crocodile  &  0.4&0.3&0.7&3&3&2e-5& 1e-3&0.5&2\\
        Squirrel &  0.5&0.3&0.7&3&3&2e-4& 1e-3&0.5&2\\
        Actor &  0.7&0.3&0.7&3&3&2e-4& 1e-3&0.9&1\\
       Roman  &  0.4&0.3&0.7&3&3&2e-4& 1e-3&0.75&2\\
        \bottomrule
    \end{tabular}}
    \label{hyper}
\end{table}

\section{More Experimental Results}
\label{hyper_experiment}

In this section, we supplement more experiments to show the effectiveness of our proposed method.
\begin{table*}[htbp]
    \centering
        \caption{ Graph classification performance comparison. We report accuracy \(\%\) for all datasets.}
    \begin{tabular}{c|ccccccc}
        \toprule
        Dataset  & Infograph & GraphMAE & GraphCL& S2GAE &GraphMAE2 &  AUG-MAE & \textbf{DGMAE}  \\
        \midrule
        IMDB-M & 49.69$\pm$0.53&51.63$\pm$0.52& 48.58$\pm$0.67& 51.79$\pm$0.36&51.80$\pm$0.60&\underline{51.80$\pm$0.86}&\textbf{51.95$\pm$0.30}\\
        PROTEIN & 74.44$\pm$0.31&75.30$\pm$0.39  &74.39$\pm$0.45& \underline{76.37$\pm$0.43}&74.86$\pm$0.34&75.83$\pm$0.24&\textbf{76.51$\pm$0.52}\\
        COLLAB &70.65$\pm$1.13& 80.32$\pm$0.46 &71.36$\pm$1.15& \underline{81.02$\pm$0.53}&77.59$\pm$0.22&80.48$\pm$0.50&\textbf{81.15$\pm$0.30}\\
        MUTAG & \underline{89.01$\pm$1.13}& 88.19$\pm$1.26 &86.80$\pm$1.34& 88.26$\pm$0.76&86.63$\pm$1.33&88.28$\pm$0.98&\textbf{89.06$\pm$0.64}\\

        \bottomrule
    \end{tabular}
    \label{table.graphclassification}
\end{table*}

\subsection{Graph Classification Performance Comparsion}
\label{grapgclasification}
Since we required a graph-level representation, the discrepancy selection module is not applied to avoid the discrepancy selection process losing some important information and considering the discrepancies of the node with each of its neighbors. As can be observed in Table \ref{table.graphclassification} although our approach considers node-level information, the discrepancy information between nodes makes the graph representation more comprehensive. The graph classification performance still achieved better performance compared to the state-of-the-art GMAE method.

\subsection{Performance on Discrepancies in Neighborhoods of Orders}
\begin{table}
\small
\renewcommand{\arraystretch}{1.2}
\renewcommand{\tabcolsep}{1.9pt}    
\label{table-operators}
\caption{The effect of the neighborhood range.}
\begin{tabular}{l|c c c c}
\toprule
Neighbor range  & Roman & Actor &Citeseer&Photo  \\
\midrule
one-hop & 74.66$\pm$0.32&36.41$\pm$0.55&73.68$\pm$0.73&93.62$\pm$0.38   \\
two-hop & 71.28$\pm$0.41&35.72$\pm$0.47&71.59$\pm$0.30&93.47$\pm$0.32\\
one-hop\&two-hop & 76.62$\pm$0.41&36.61$\pm$0.74&73.70$\pm$0.36&93.75$\pm$0.30\\ 
\bottomrule
\end{tabular}
\label{neighbor_range}
\end{table}

To investigate the impact of varying neighbor orders on discrepancy reconstruction, we evaluated the performance of nodes with first-order neighbors, second-order neighbors, and the combination of the discrepancies between first-order and second-order neighbors. Table \ref{neighbor_range} shows that the discrepancies between first-order neighbors are more significant and effective than those between second-order neighbors. This is because first-order neighbors represent the original structure of nodes and neighborhoods. By learning the discrepancies between nodes and their first-order neighbors, we can prevent nodes and neighborhoods from becoming overly similar and obtain a more distinguishable node representation, and second-order neighbours provide higher-order discrepancy information.

\subsection{Performance on Synthetic Graphs}
\begin{figure}
\begin{minipage}[b]{0.45\columnwidth}
    \includegraphics[width=\columnwidth]{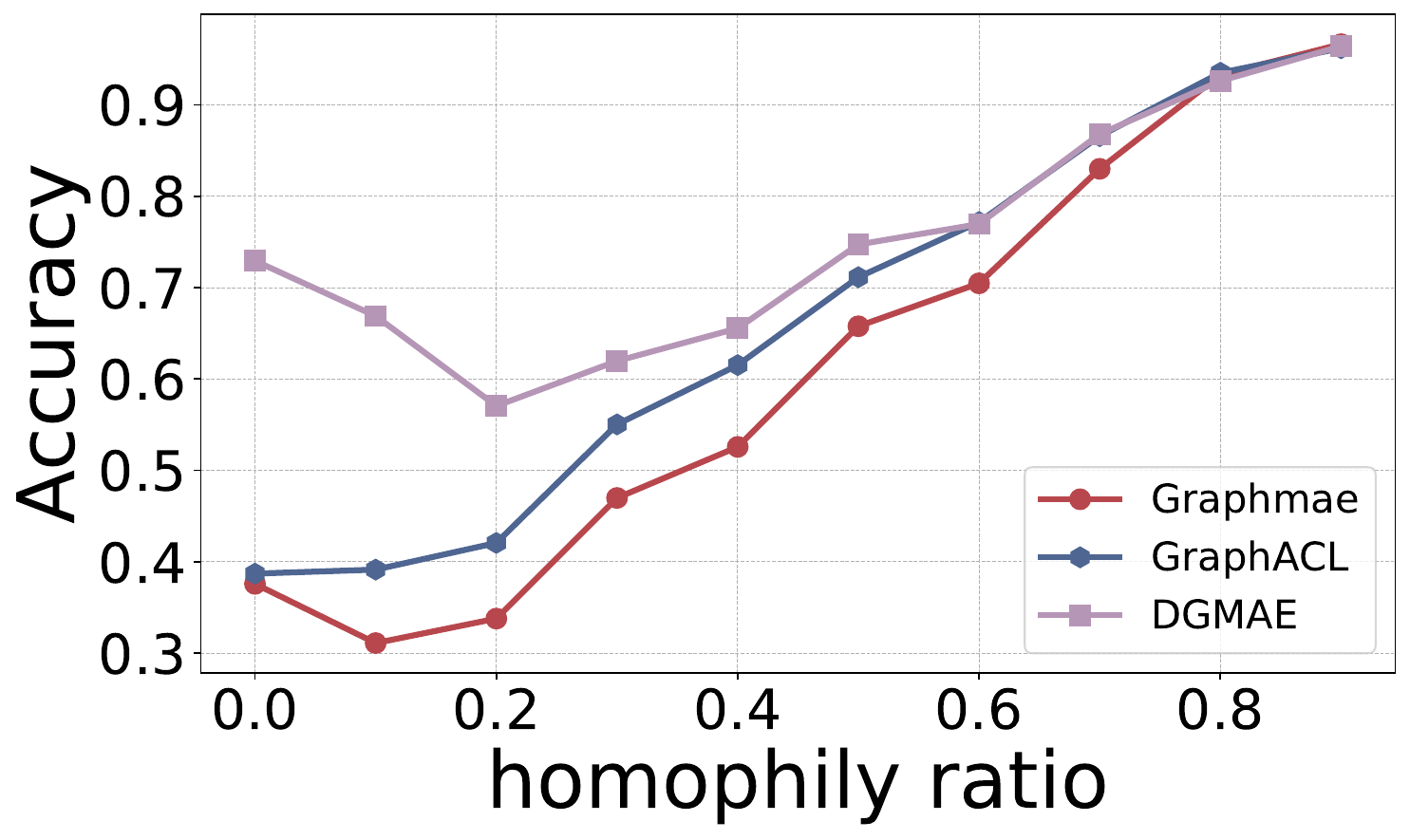}
  \caption{The performance on the syn-cora dataset of different homophily ratios.}
    \label{syngraph}
    \end{minipage}
    \hspace{0.35cm}
      \begin{minipage}[b]{0.45\columnwidth}
   \includegraphics[width=\columnwidth]{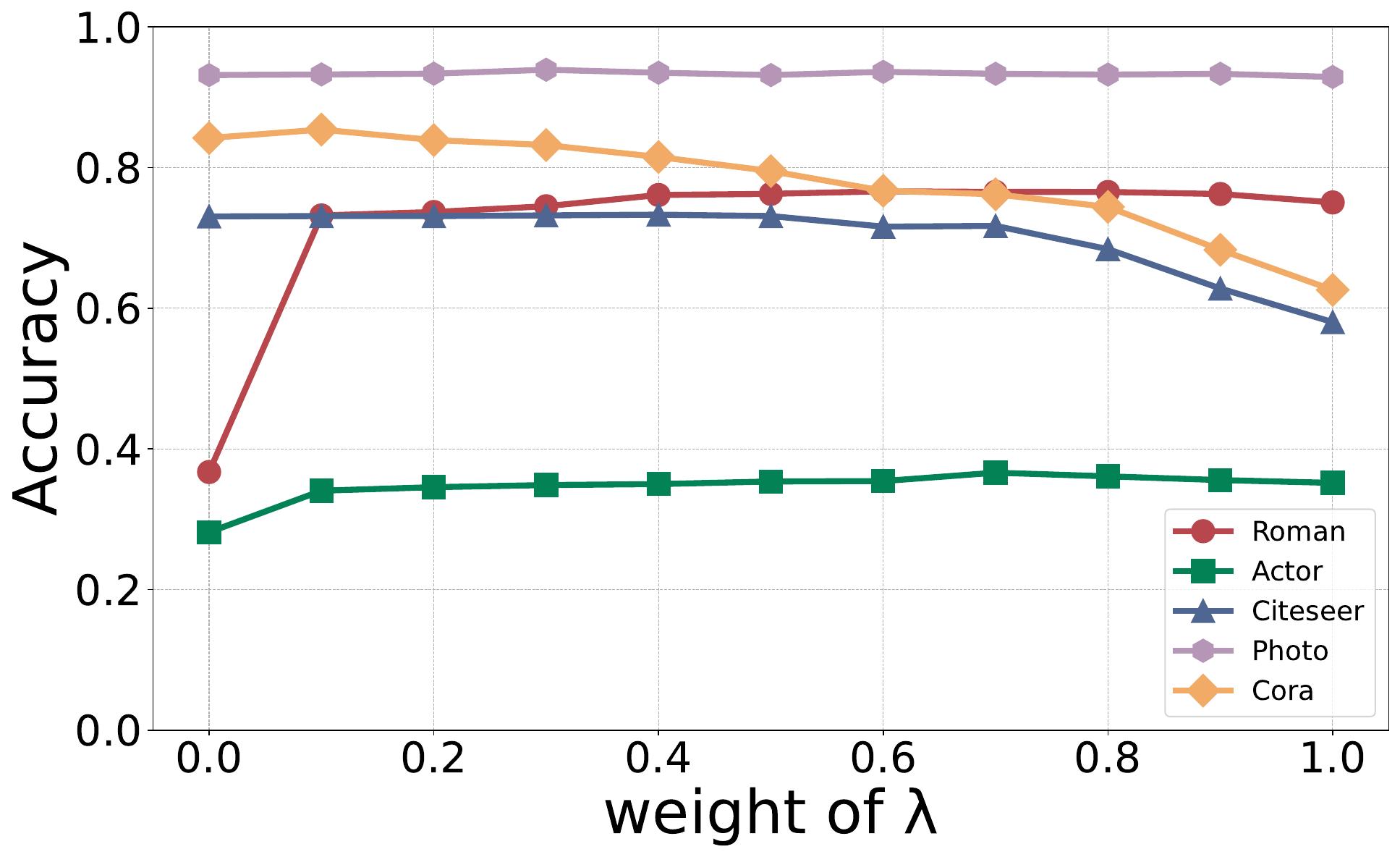}
    \caption{The effect of discrepancy reconstruct weight $\lambda$.}
    \label{disweight}
   \end{minipage}
    \Description{..}
\end{figure}
To investigate the effect under different levels of homophily, we tested the variation curves of several self-supervised methods on a synthetic graph~\cite{h2gnn}. The results are shown in Figure \ref{syngraph}, which shows that our method significantly outperforms other self-supervised methods under strong heterophily(homophily ratio<0.3), especially when the homophily is 0, the edge distribution tends to be randomly distributed, it becomes difficult for GraphACL to predict the neighborhood distribution, and GraphMAE suffers from the lack of sufficient contextual information to reconstruct the mask nodes, which causes both methods to be ineffective under low homophily with average results. Compared to them, DGMAE is still capable of obtaining distinguishable node representations under low homogeneity by learning the discrepancy information between nodes. Our method also performs well at high homophily ratios. This shows that our method applies to learning in different heterophilic environments.

\subsection{Sensitivity Analysis of More Hyperparameters}
\subsubsection{Discrepancy Weight $\lambda$}
To investigate the impact of different weights on various datasets, we examined the effect of different $\lambda$ on homophilic and heterophilic graphs, as shown in Figure \ref{disweight}. The results indicate that the classification performance of nodes on homophilic graphs decreases as the weight of the loss exceeds $0.1$. Conversely, the performance of heterophilic graphs gradually improves before slightly decreasing. Therefore, too large $\lambda$ will instead force the model to learn discrepancies among similar nodes, leading to poor performance.

\begin{figure}[ht]
  \centering
   \begin{minipage}[b]{\columnwidth}
    \setlength{\abovecaptionskip}{0.cm}
    \subfigure[Node Classification]{\includegraphics[width=0.48\columnwidth]{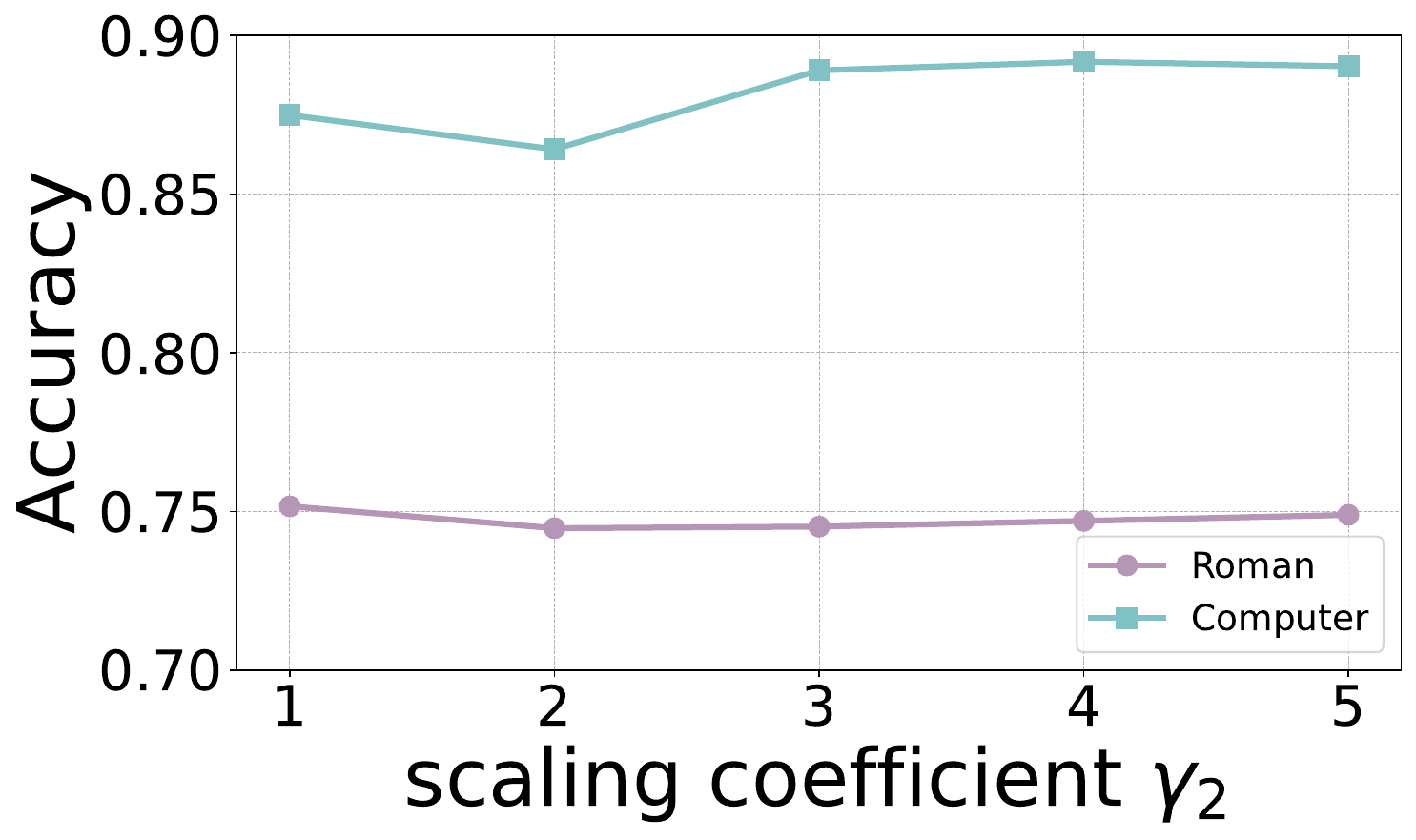}}
  \subfigure[Graph Classification]{\includegraphics[width=0.48\columnwidth]{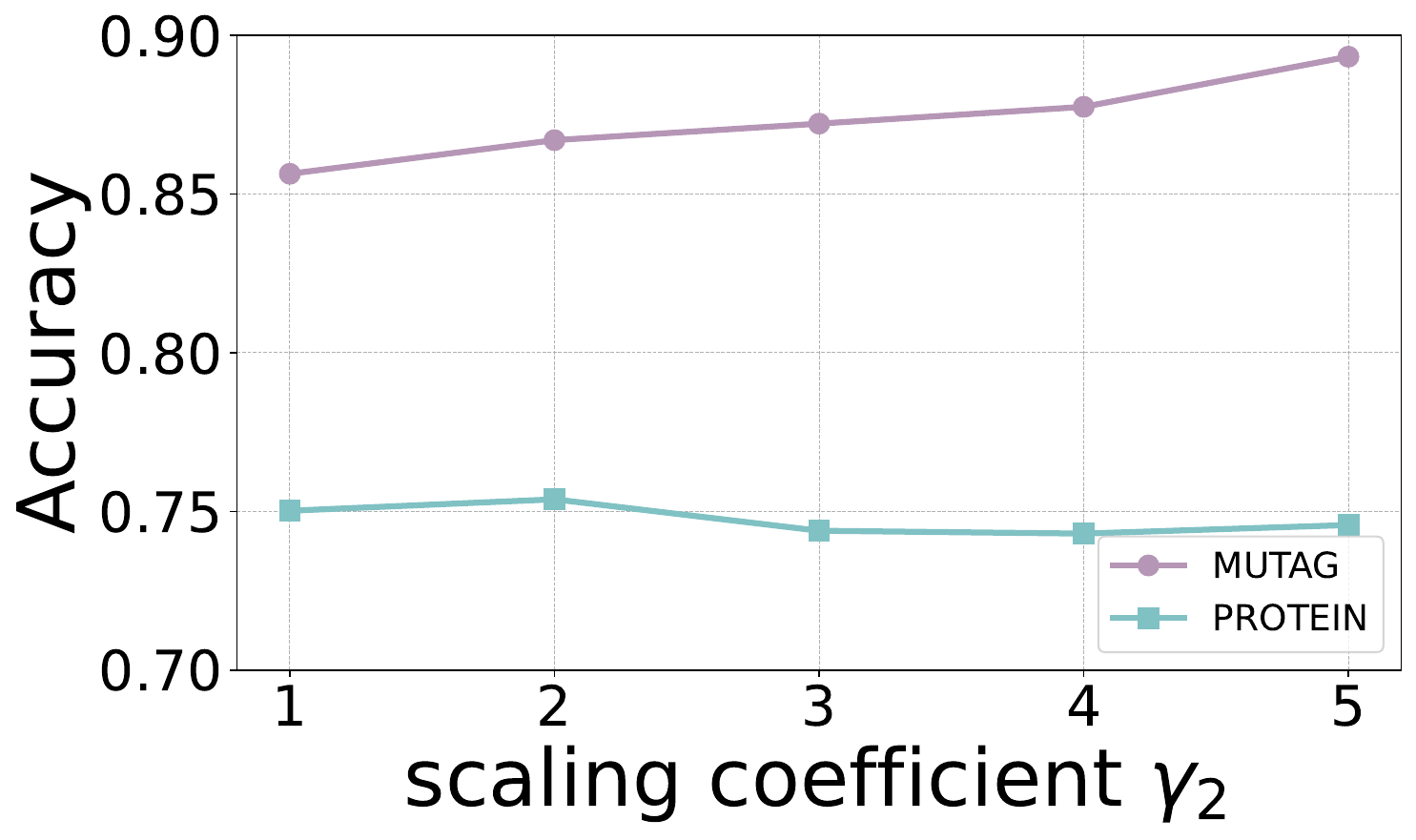}}
    \caption{The effect on node classification and graph classification of scaling coefficient $\gamma_2$.}
    \label{scaling}
   \end{minipage}
      \Description{..}
\end{figure}

\begin{figure}[h]
   \centering
  \subfigure[Cora]{
      \includegraphics[width=0.48\columnwidth]{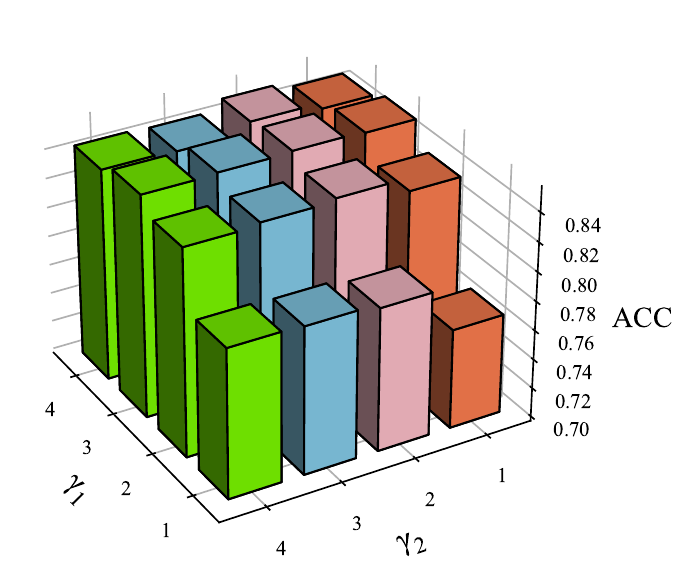}}
  \subfigure[Roman]{
      \includegraphics[width=0.48\columnwidth]{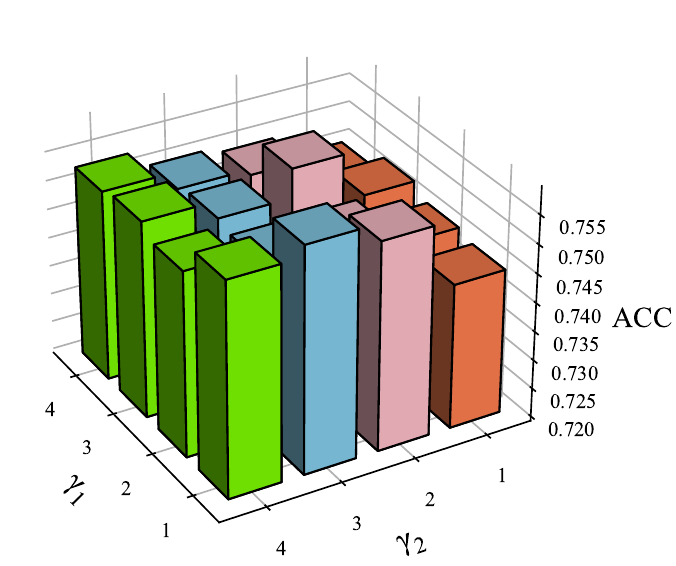}}
    \caption{The effect of $\gamma_1$ and $\gamma_2$ on homophilic graph Cora and heterophilic graph Roman.}
    \label{y1y2}
\end{figure}
\subsubsection{Scaling Coefficient $\gamma_1$ and $\gamma_2$}
$\gamma_1$ and $\gamma_2$  are used to scale down the weights of the samples in feature reconstruction loss and discrepancy reconstruction loss, respectively. To explore the effect of these two parameters on the learned representation at different values, we investigated the joint effect of these two parameters on the performance of the homophilous graph Cora and heterophilous graph Roman. As shown in Figure.\ref{y1y2},  for the homophilic graph Cora, raising $\gamma_1$ can significantly improve the classification performance, which requires the model to focus on hard-to-reconstruct samples in the feature reconstruction, while changing  $\gamma_2$ has relatively little impact on the homography due to the small feature differences between nodes and their neighbors. For the heterophilic graphs Roman, the effects of $\gamma_1$ and $\gamma_2$ on classification performance display opposite trends, and due to the large differences between node features in heteroscedastic graphs, difference reconstruction plays a more important role in this process.

\subsubsection{Scaling Coefficient $\gamma_2$}
In addition, the impact of the scaling coefficient $\gamma_2$ on the discrepancy loss was investigated in Figure \ref{scaling}. An increase in the scaling coefficient is crucial for enhanced performance.  In the case of the Roman dataset, given that the connected nodes are different classes with a greater degree of discrepancy, an increase in the gamma may result in some discrepancies being overlooked. Conversely, in other datasets, a larger value will prompt the model to prioritize the mask nodes with a high reconstruction loss, thus improving the model's performance. 



\begin{figure}[ht]
  \centering
  \subfigure[Citeseer]{\includegraphics[width=0.445\columnwidth]{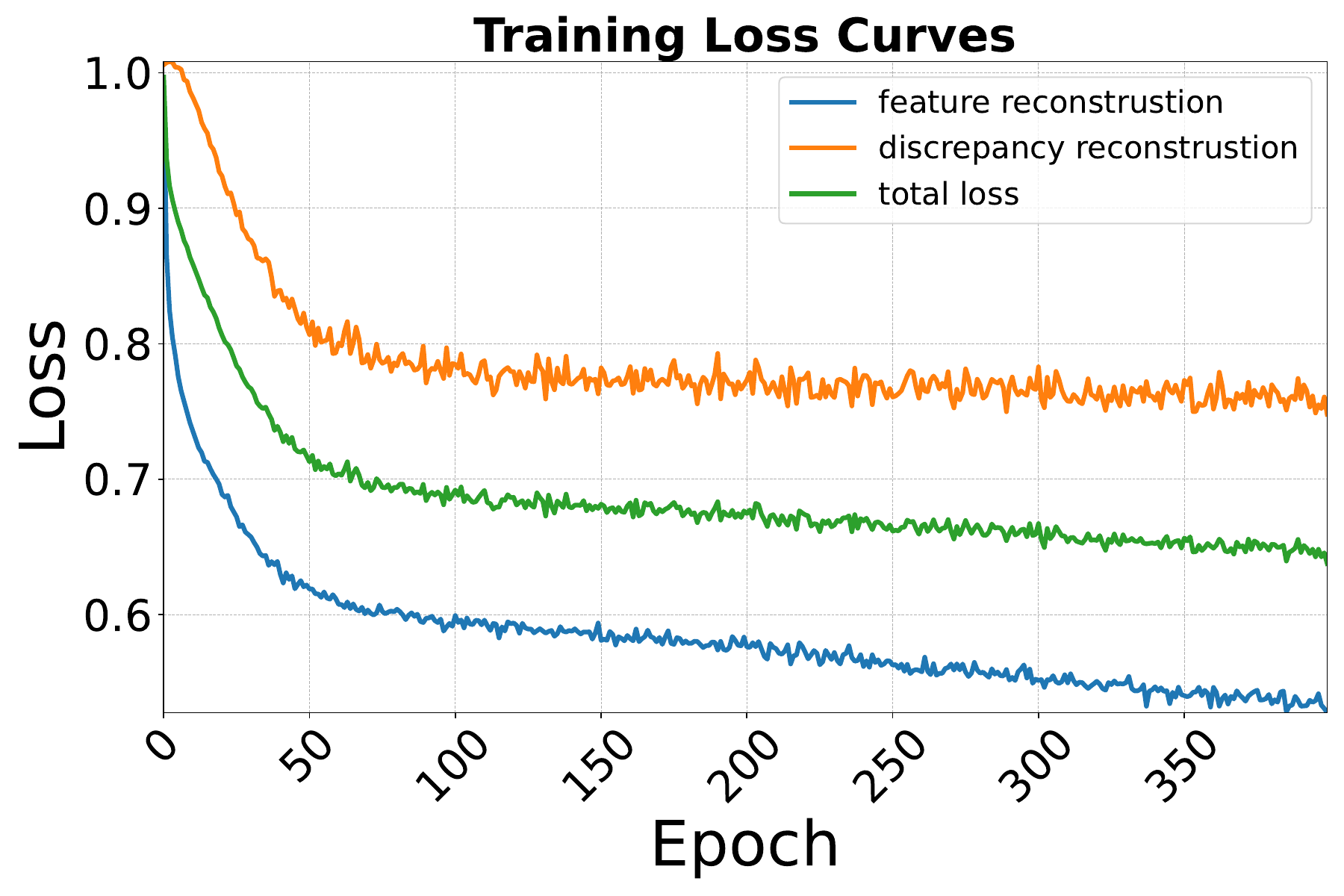}}
  \subfigure[Actor]{\includegraphics[width=0.445\columnwidth]{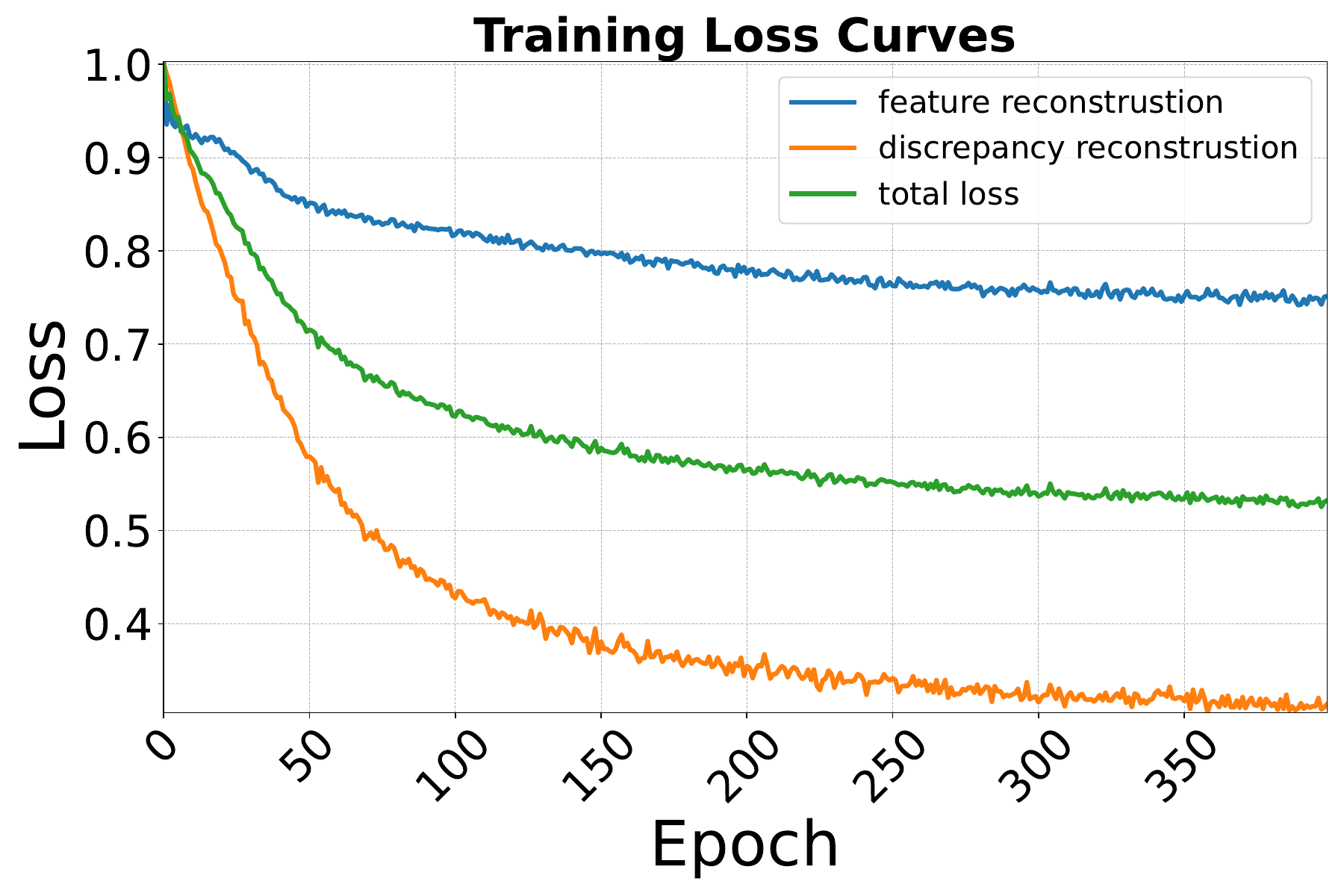}}
  \caption{The training cruves of different loss in homophilic graph Citeseer and heterphilic graph Actor.}
  \label{train.cruves}
\end{figure}

\begin{table*}

    \centering
\caption{Experimental results on 5 larger heterophilic graphs. Accuracy is reported for Amazon-ratings and Arxiv-year, and ROC AUC is reported for Minesweeper, Tolokers, and Questions~\cite{roman}. OOMdenotes"out of memory". The best and second-best results are highlighted in bold and
underlined, respectively}
    \begin{tabular}{c|ccccccc}
        \toprule
        Method  & Minesweeper & Question & Tolokers &  Amazon-ratings & Arxiv-year & Twitch-gamer & Penn94\\
        \midrule
        CCA-SSG &  70.20$\pm$0.38 & \underline{75.23$\pm$0.56} &  75.02$\pm$0.47 &41.12$\pm$0.38  & 39.51$\pm$0.45 &60.26$\pm$0.45&63.46$\pm$0.36\\
        GCA   & 72.83$\pm$0.56 & 74.54$\pm$0.32 & 78.20$\pm$0.42 &\underline{42.30$\pm$0.12} & \underline{42.52$\pm$0.25} &OOM & 59.67±0.48\\
        
        GREET   & \underline{81.23$\pm$0.78} & 73.23$\pm$0.78 & \underline{80.90$\pm$0.45} & 41.52$\pm$0.64 & OOM &OOM &OOM\\
        \textbf{DGMAE} & \textbf{90.72$\pm$0.48} & \textbf{77.93$\pm$0.98} & \textbf{83.48$\pm$0.57} & \textbf{45.60$\pm$0.69} & \textbf{43.41$\pm$0.64} &\textbf{63.57$\pm$0.25} & \textbf{76.69$\pm$0.51}\\  
        \bottomrule
    \end{tabular}
       \label{large_heter}
\end{table*}

\subsection{The Effect of Contextual Discrepancy}
 To explore the contribution of contextual discrepancy in discrepancy reconstruction, we replace the discrepancy between nodes and their neighbors with pairwise discrepancy of nodes, shift the discrepancy reconstruction from the node perspective to the edge perspective, and consider only the feature discrepancy between neighboring pairs. We compute the feature discrepancy $z_i-z_j$ of two nodes connected to each edge, and the reconstruction target $x_i-x_j$ to perform the discrepancy reconstruction of pairwise relations. Table ~\ref{tab:discrepancy-transposed} shows that focusing on neighborhood discrepancy is more effective, which is because contextual discrepancy preserves the unique information experiments that distinguish nodes from their neighbors by reducing the information that nodes have in common with their neighbors. This also proves the effectiveness of contextual differences in discrepancy reconstruction.
 
\subsection{Node Classification on the Larger Heterophilic Graph}
\label{newheter}
We also conducted experiments on 5 larger heterogeneous graphs, and as shown in the Table.\ref{large_heter} our method remains valid on the larger heterophilic graph. DGMAE shows significant performance improvement compared to GREET~\cite{greet}, especially on Minesweeper by $9.49\%$ and $13.23\%$ improvement over the suboptimal method on the Penn94. The excellent performance of DGMAE suggests that it is beneficial to capture information about the discrepancy between features.

\subsection{The Stability and Convergence of DGMAE}
We give training curves for feature reconstruction loss, discrepancy reconstruction loss and total loss. As shown in Figure.\ref{train.cruves}, the feature reconstruction loss decreases faster than the discrepancy reconstruction loss for the homophilic graph Citeseer, while the opposite trend is shown in the heterophilic graph Actor. This is because for heterophilic graphs it is difficult to recover features by dissimilar neighbors, whereas in homophilic graphs, the discrepancy between neighboring nodes are smaller and it is more difficult to learn the discrepancy. We can see that all three loss curves show a trend of decreasing and stabilizing loss with increasing Epoch.

\begin{figure}[t]
\centering
\includegraphics[width=0.8\columnwidth]{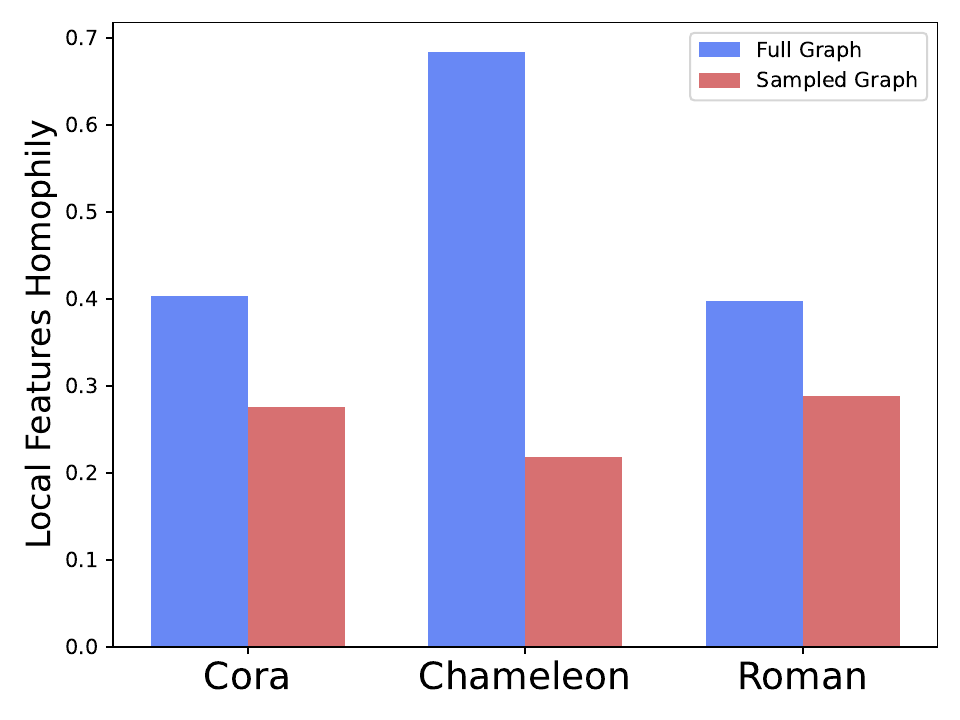}
\caption{The local feature homophily on full graph and sampled graph of different datasets.}
\label{feat_homo}
\Description{..}
\end{figure}

\subsection{Effect of discrepancy selection on homophily}
To prove that the edge selection strategy captures the discrepancy between features, We used the local homophily metric~\cite{lsgnn} to measure the homophily of the graphs. It measures feature homophily at the node level based on the hypothesis that nodes with similar features are likely to belong to the same class. The definition is as follows:
\begin{equation}
h(G, X) = \frac{1}{|V|}\sum \limits_{i \in \mathcal{V}}\frac{1}{d_i}\sum\limits_{v \in \mathcal{N}(i)}\frac{X_i X_j}
{||X_i||||X_j||}.
\end{equation}

We calculated the homophily of the full and sampled graphs under different datasets, and the results are shown in Figure.\ref{feat_homo}, where the sampled graphs have lower local homophily compared to the original graphs, which implies that there are higher feature discrepancies between nodes and their neighbors.

\begin{table}[h]
\caption{Performance on different discrepancy}
\centering
\begin{tabular}{c|cccc}
\hline
 & contextual discrepancy & pair discrepancy & & \\
\hline
Actor & \textbf{36.61±0.74} & 33.59±0.39 & & \\
Squirrel & \textbf{72.47±1.77} & 60.18±1.64 & & \\
Chameleon & \textbf{75.50±1.17} & 74.23±1.17 & & \\
Roman & \textbf{76.62±0.41} & 72.18±0.44 & & \\
\hline
\end{tabular}
\label{tab:discrepancy-transposed}
\end{table}

\begin{table}
\caption{Memory, training time, and parameters of different models.}
\begin{tabular}{l|c c c}
\toprule
\textbf{Metric} & GREET  & Bandana &  \textbf{DGMAE}\\
\midrule
Memory(M) & 4820&	2877&	\textbf{1416} \\
Params (M) & 2.23 &  1.89 & \textbf{1.56} \\
Time (s) & 28.75 & 3.29 & \textbf{1.50} \\
\bottomrule
\end{tabular}
\label{complexity}
\end{table}

\subsection{Complexity Analysis}
The time complexity of DGMAE consists of four components:  encoding, discrepancy selection, decoding, and loss computation. We use $\mathcal{|V|}$ to denote the node set and $\mathcal{|E|}$ to denote the edge set.  The complexity of the encoder is $O(|\mathcal{V}|dd' + |\mathcal{E}|d')l$, $l$ is the number of the hidden layer, and the complexity of the decoder is $O(|\mathcal{V}|d'd)$, $d'$ denotes the hidden layer dimension. For the discrepancy selection,  due to the sampling weights of the edges being shared with GAT and do not need to be recomputed, the time complexity is $O(|\mathcal{E}|)$. The total number of nodes in the loss calculation process is the sum of masked and unmasked nodes, so the time complexity of the overall loss is $O(|\mathcal{V}|d)$.

We take the Roman dataset as an example and give the number of parameters and the training time for one iteration achieved by the different models, for which the number of network layers and feature dimensions of the encoder are set to be the same. The results are shown in the Table.\ref{complexity}, it can be seen that the number of parameters of our method compared to other methods does not increase the cost too much, and in terms of time complexity, DGMAE one iteration of training time is the shortest, while GREET needs to spend a lot of time for each training. This is because GREET needs to additionally train an edge discriminator, compared to the time complexity and space complexity of our method, which is acceptable.

\section{Analysis of the Effectiveness of Discrepancy Reconstruction Loss}
\label{loss analysis}
We analyze why discrepancy reconstruction losses are effective. Discrepancy reconstruction could be regarded as the process of decoupling the node's own unique representation and the neighbor's consistent representation. Given a node $v_i$ and a neighbor node $v_j$, consistent representation indicates the common information of nodes $v_i$ and $v_j$, and unique representation indicates the information of nodes specific to their neighbors. 

We use the $x_i^D$ obtained from Eq.\ref{Eq9} as a supervised signal for feature decoupling. $x_i^D$ is the feature obtained after Laplace sharpening, which amplifies the difference between a node's and neighbor's features. Next, we further analyze the discrepancy loss:
\begin{equation}
\mathop{min} \quad\mathcal{L}_d  = \frac{1}{\hat{V}} \sum \limits_{i \notin \hat{|V|}} (1- 
⟨\mathbf{z_i^D},\mathbf{x_i^D}⟩)^{\gamma_2} \iff \mathop{max} \quad ⟨\mathbf{z_i^D},\mathbf{x_i^D}⟩\\
\end{equation}
Let $\mathbf{x_i^D}=\sum \limits_{\mathcal{j} \in \mathcal{N}(i)'}(\mathbf{x_i}-\mathbf{x_j})$, $\mathcal{N}(i)'$ denotes the neighborhood of node $v_i$ after adaptive sampling. Thus the minimizing loss $\mathcal{L}_d$ can be expressed as:
\begin{equation}
    \begin{aligned}
\mathop{max} \quad⟨\mathbf{z_i^D} ,\mathbf{x_i^D}⟩ 
&=⟨\mathbf{z_i^D} ,\sum_{j \in \mathcal{N}(i)'} (\mathbf{x_i}-\mathbf{x_j})⟩\\
&=⟨\mathbf{z^D},\mathbf{x_i}⟩-\sum_{j \in \mathcal{N}(i)'} ⟨\mathbf{z^D},\mathbf{x_j}⟩ \quad (\mathbf{x}\  is \ normalized)\\
&=-\frac{1}{2}\mathop{\underline{||\mathbf{z^D_i}-\mathbf{x_i}||}^2}\limits_{pull\ closer} +\sum \limits_{j \in \mathcal{N}(i)'}\frac{1}{2}\mathop{\underline{||\mathbf{z^D_i}-\mathbf{x_j}||}^2}\limits_{push\ away}
\end{aligned}
\end{equation}

In the process of optimizing the loss, the original feature information is preserved by minimizing the distance between  $\mathbf{z_i^D}$ and $\mathbf{x_i}$.  $\mathbf{x_j}$  is the original feature of the neighboring node, maximizing the distance between the $\mathbf{z_i^D}$ and $\mathbf{x_j}$. In this process,  $\mathbf{z_i^D}$ maintains the discrepancy between the original features to avoid convergence of node representations.

Therefore, discrepancy reconstruction is understood as the process of decoupling the node's unique representation and consistent representation. Consistent representation denotes the common information of nodes $v_i$ and $v_j$, and unique representation denotes the information of node  $v_i$ specific to neighbors. 

Eq.11, $\mathbf{z_i}= \mathbf{z_i^D} +\mathbf{\hat{z}_i}$,   $\mathbf{\hat{z}_i}$ is obtained from the GAT decoder, which recovers the features of the masked node with the common information between the neighbors and the masked node, thus $\mathbf{\hat{z}_i}$ denoting the consistent representation. $\mathbf{z_i^D}$ is the unique representation.

\section{Limitations}
\label{limited}
The core of our approach is to guide low-dimensional node representations based on raw feature discrepancies. However, in practice, some graph data may contain only graph structures. For this type of graph data, our approach may not be applicable. Moreover, forcing the model to learn the discrepancies between similar nodes may lead to misclassification of similar nodes and obtaining wrong decision boundaries. Discrenpancy information is not always valid and may come from noisy data, and it is also important to know how to distinguish beneficial difference information. Except for the discrepancies in node features, homophilic and heterophilic nodes also differ in neighborhood structure, and further mining the impact of discrepancy information from different perspectives can help unify the research related to homophilic and heterophilic graphs.

\end{document}